# Exploiting Model Equivalences for Solving Interactive Dynamic Influence Diagrams

**Yifeng Zeng**                                                    YFZENG@CS.AAU.DK
*Dept. of Computer Science*
*Aalborg University*
*DK-9220 Aalborg, Denmark*

**Prashant Doshi**                                                 PDOSHI@CS.UGA.EDU
*Dept. of Computer Science*
*University of Georgia*
*Athens, GA 30602, U.S.A.*

## Abstract

We focus on the problem of sequential decision making in partially observable environments shared with other agents of uncertain types having similar or conflicting objectives. This problem has been previously formalized by multiple frameworks one of which is the *interactive dynamic influence diagram (I-DID)*, which generalizes the well-known influence diagram to the multiagent setting. I-DIDs are graphical models and may be used to compute the policy of an agent given its belief over the physical state and others' models, which changes as the agent acts and observes in the multiagent setting.

As we may expect, solving I-DIDs is computationally hard. This is predominantly due to the large space of candidate models ascribed to the other agents and its exponential growth over time. We present two methods for reducing the size of the model space and stemming its exponential growth. Both these methods involve aggregating individual models into equivalence classes. Our first method groups together *behaviorally equivalent* models and selects only those models for updating which will result in predictive behaviors that are distinct from others in the updated model space. The second method further compacts the model space by focusing on portions of the behavioral predictions. Specifically, we cluster *actionally equivalent* models that prescribe identical actions at a single time step. Exactly identifying the equivalences would require us to solve all models in the initial set. We avoid this by selectively solving some of the models, thereby introducing an approximation. We discuss the error introduced by the approximation, and empirically demonstrate the improved efficiency in solving I-DIDs due to the equivalences.

## 1. Introduction

Sequential decision making (planning) is a key tenet of agent autonomy. Decision making becomes complicated due to actions that are nondeterministic and a physical environment that is often only partially observable. The complexity increases exponentially in the presence of other agents who are themselves acting and observing, and whose actions impact the subject agent. Multiple related frameworks formalize the general problem of decision making in uncertain settings shared with other sophisticated agents who may have similar or conflicting objectives. One of these frameworks is the interactive partially observable Markov decision process (I-POMDP) (Gmytrasiewicz & Doshi, 2005), which generalizes POMDPs (Smallwood & Sondik, 1973; Kaelbling, Littman, & Cassandra, 1998) to multiagent settings; another framework is the *interactive dynamic influ-*

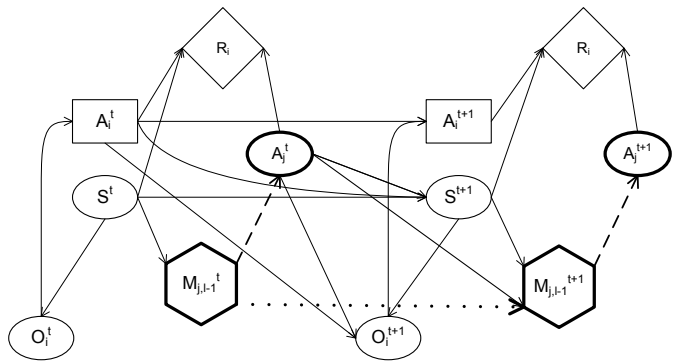

Figure 1: A two time-slice I-DID for agent $i$ modeling another agent $j$. I-DIDs allow representing models in a model node (hexagon) and their update over time using the dotted model update link. Predictions about the other agent's behavior from the models are represented using a dashed policy link.

*ence diagram (I-DID)* (Doshi, Zeng, & Chen, 2009). In cooperative settings, the decentralized POMDP (Bernstein, Givan, Immerman, & Zilberstein, 2002) framework models multiagent decision making.

I-DIDs are graphical models for sequential decision making in uncertain multiagent settings. They concisely represent the problem of how an agent should act in an uncertain environment shared with others who may act simultaneously in sophisticated ways. I-DIDs may be viewed as graphical counterparts of I-POMDPs which adopt an enumerative representation of the decision-making problem. I-DIDs generalize dynamic influence diagrams (DID) (Tatman & Shachter, 1990) to multiagent settings analogously to the way that I-POMDPs generalize POMDPs. Importantly, I-DIDs have the advantage of a representation that explicates the embedded domain structure by decomposing the state space into variables and relationships between the variables. Not only is this representation more intuitive to use, it translates into computational benefits when compared to the enumerative representation as used in I-POMDPs (Doshi et al., 2009).

Following the paradigm of graphical models, I-DIDs compactly represent the decision problem by mapping various variables into chance, decision and utility nodes, and denoting the dependencies between variables using directed arcs between the corresponding nodes. They extend DIDs by introducing a special *model node* whose values are the possible models of the other agent. These models may themselves be represented using I-DIDs leading to nested modeling. Both other agents' models and the original agent's beliefs over these models are updated over time using a special *model update* link that connects the model nodes between time steps. Solution to the I-DID is a policy that prescribes what the agent should do over time, given its beliefs over the physical state and others' models. Consequently, I-DIDs may be used to compute the policy of an agent *online* – given an initial belief of the agent – as the agent acts and observes in a setting that is populated by other interacting agents. We show a generic I-DID in Fig. 1 and provide more details in Section 3.

As we may expect, solving I-DIDs is computationally very hard. In particular, they acutely suffer from the curses of dimensionality and history (Pineau, Gordon, & Thrun, 2006). This is because the state space in I-DIDs includes the models of other agents in addition to the traditional physical states. As the agents act, observe and update beliefs, I-DIDs must track the evolution of the models over time. Theoretically, the number of candidate models grows exponentially over time. Thus, I-DIDs not only suffer from the curse of history that afflicts the modeling agent, but also from that exhibited by the modeled agents. This is further complicated by the nested nature of the state space.

Consequently, exact solutions of I-DIDs are infeasible for all but the simple problems and ways of mitigating the computational intractability are critically needed. Because the complexity is predominantly due to the candidate models, we focus on principled reductions of the model space while avoiding significant losses in the optimality of the decision maker. Our first approach builds upon the idea of grouping together *behaviorally equivalent* (BE) models (Rathnasabapathy, Doshi, & Gmytrasiewicz, 2006; Pynadath & Marsella, 2007). These are models whose behavioral predictions for the modeled agent(s) are identical. Because the solution of the subject agent's I-DID is affected only by the predicted behavior of the other agent regardless of the description of the ascribed model, we may consider a single representative from each BE class without affecting the optimality of the solution. Identifying BE models requires solving the individual models. We reduce the exponential growth in the model space by discriminatively updating models. Specifically, at each time step, we select only those models for updating which will result in predictive behaviors that are distinct from others in the updated model space. In other words, models that on update would result in predictions which are identical to those of existing models are not selected for updating. For these models, we simply transfer their revised probability masses to the existing BE models. Thus, we avoid generating all possible updated models and subsequently reducing them. Rather, we generate a *minimal set* of models at each time step.

Restricting the updated models to the exact minimal set would require solving all the models that are considered initially. Exploiting the notion that models whose beliefs are spatially close tend to be BE, we solve those models only whose beliefs are not $\epsilon$-close to a representative. We theoretically analyze the error introduced by this approach in the optimality of the solution. Importantly, we experimentally evaluate our approach on I-DIDs formulated for multiple problem domains having two agents, and show approximately an order of magnitude improvement in performance in comparison to the previous clustering approach (Zeng, Doshi, & Chen, 2007), with a comparable loss in optimality. One of these problem domains is the *Georgia testbed for autonomous control of vehicles (GaTAC)* (Doshi & Sonu, 2010), which facilitates scalable and realistic problem domains pertaining to autonomous control of unmanned agents such as uninhabited aerial vehicles (UAV). GaTAC provides a low-cost, open-source and flexible environment for realistically simulating the problem domains and evaluating solutions produced by multiagent decision-making algorithms.

We further compact the space of models in the model node by observing that behaviorally distinct models may prescribe identical actions at a single time step. We may then group together these models into a single equivalence class. In comparison to BE, the definition of our equivalence class is different: it includes those models whose prescribed action for the *particular* time step is the same, and we call it *action equivalence* (AE). Since there are typically additional models than the BE ones that prescribe identical actions at a time step, an AE class often includes many more models. Consequently, the model space is partitioned into lesser number of classes than previously and is bounded by the number of actions of the other agent.

Unlike the update of BE classes, given the action and an observation AE classes do not update deterministically. We show how we may compute the probability with which an equivalence class is updated to another class in the next time step. Although, in general, grouping AE models introduces an approximation, we derive conditions under which AE model grouping preserves optimality of the solution. We demonstrate the performance of our approach on multiple two-agent problem domains including in GaTAC and show significant time savings in comparison to previous approaches.

To summarize, the main contributions of this article are new approaches that group equivalent models more efficiently leading to improved scalability in solving I-DIDs. Our first method reduces the exponential growth of the model space by discriminatively updating models thereby generating a behaviorally minimal set in the next time step that is characterized by the absence of BE models. The second method adopts a relaxed grouping of models that prescribe identical actions for that particular time step. Grouping AE models leads to equivalence classes that often include many models in addition to those that are BE. We augment both these methods with an approximation that avoids solving all of the initial models, and demonstrate much improved scalability in experiments.

The remainder of this article is structured as follows. In Section 2, we discuss previous work related to this article. In Section 3, we briefly review the graphical model of I-DID as well as its solution based on BE. In Section 4, we show how we may discriminatively update models in order to facilitate behaviorally-distinct models at subsequent time steps. We introduce an approximation, and discuss the associated computational savings and error. We introduce the approach of further grouping models based on actions, in Section 5. All approaches for solving I-DIDs are empirically evaluated along different dimensions in Section 6. We conclude this article with a discussion of the framework and the solution approaches including extensions to N > 2 agent interactions, and some limitations, in Section 7. The Appendices contain proofs of propositions mentioned elsewhere, and detailed descriptions and I-DID representations of the problem domains used in the evaluation.

## 2. Related Work

Suryadi and Gmytrasiewicz (1999) in an early piece of related work, proposed modeling other agents using IDs. The approach proposed ways to modify the IDs to better reflect the observed behavior. However, unlike I-DIDs, other agents did not model the original agent and the distribution over the models was not updated based on the actions and observations.

As detailed by Doshi et al. (2009), I-DIDs contribute to an emerging and promising line of research on graphical models for multiagent decision making. This includes multiagent influence diagrams (MAID) (Koller & Milch, 2001), network of influence diagrams (NID) (Gal & Pfeffer, 2008), and more recently, limited memory influence diagram based players (Madsen & Jensen, 2008). While MAIDs adopt an external perspective of the interaction, exploiting the conditional independence between effects of actions to compute the Nash equilibrium strategy for all agents involved in the interaction, I-DIDs offer a subjective perspective to the interaction, computing the best-response policy as opposed to a policy in equilibrium. The latter may not account for other agent's behaviors outside the equilibrium and multiple equilibria may exist. Furthermore, both MAID and NID formalisms focus on a static, single-shot interaction. In contrast, I-DIDs offer solutions over extended time interactions, where agents act and update their beliefs over others' models which are themselves dynamic.

While I-DIDs closely relate to the previously mentioned ID-based graphical models, another significant class of graphical models compactly represents the joint behavior as a graphical game (Kearns, Littman, & Singh, 2001). It models the agents as graph vertices and an interaction in payoff between two agents using an edge, with the objective of finding a joint distribution over agents' actions possibly in equilibrium. More recently, graphical multiagent models (Duong, Wellman,

& Singh, 2008) enhance graphical games by allowing beliefs over agent behaviors formed from different knowledge sources, and conditioning agent behaviors on abstracted history if the game is dynamic (Duong, Wellman, Singh, & Vorobeychik, 2010).

As we mentioned previously, a dominating cause of the complexity of I-DIDs is the exponential growth in the candidate models over time. Using the insight that models (with identical capabilities and preferences) whose beliefs are spatially close are likely to be BE, Zeng and Doshi (2007) utilized a $k$-means approach to cluster models together and select $K$ models closest to the means of the clusters in the model node at each time step. This approach facilitates the consideration of a fixed number of models at each time. However, the approach first generates all possible models before reducing the model space at each time step, thereby not reducing the memory required. Further, it utilizes an iterative and often time-consuming $k$-means clustering method.

The concept of BE of models was proposed and initially used for solving I-POMDPs (Rathnasabapathy et al., 2006), and discussed generally by Pynadath and Marsella (2007). We contextualize BE within the framework of I-DIDs and seek further extensions. A somewhat related notion is that of state equivalence introduced by Givan et al. (2003) where the equivalence concept is exploited to factorize MDPs and gain computational benefit. Along this direction, another type of equivalence in probabilistic frameworks such as MDPs and POMDPs, sometimes also called BE, is bisimulation (Milner, 1980; Givan et al., 2003; Castro, Panangaden, & Precup, 2009). Two states are bisimilar if any action from these states leads to identical immediate reward and the states transition with the same probability to equivalence classes of states. While bisimulation is a test within a model given just its definition, BE in multiagent systems is defined and used differently: as a way of comparing between models using their solutions. Interestingly, both these concepts are ultimately useful for model minimization.

Other frameworks for modeling the multiagent decision-making problem exist. Most notable among them is the decentralized POMDP (Bernstein et al., 2002). This framework is suitable for cooperative settings only and focuses on computing the joint solution for all agents in the team. Seuken and Zilberstein (2008) provide a comprehensive survey of approaches related to decentralized POMDPs; we emphasize a few that exploit clustering. Emery-Montemerlo et al. (2005) propose iteratively merging action-observation histories of agents that lead to a small worst-case expected loss. While this clustering could be lossy, Oliehoek et al. (2009) losslessly cluster histories that exhibit probabilistic equivalence. Such histories generate an identical distribution over the histories of the other agents and lead to the same joint belief state. While we utilize BE to losslessly cluster the models of the other agent, we note that BE models when combined with the subject agent's policy induce identical distributions over the subject agent's action-observation history. More recently, Witwicki and Durfee (2010) use influence-based abstraction in order to limit an agent's belief to the other agent's relevant information by focusing on mutually-modeled features only.

Our agent models are analogous to types in game theory (Harsanyi, 1967), which are defined as attribute vectors that encompass all of an agent's private information. In this context, Dekel et al. (2006) define a strategic topology on universal type spaces (Mertens & Zamir, 1985; Brandenburger & Dekel, 1993) under which two types are close if their strategic behavior is similar in all strategic situations. While Dekel et al. focus on a theoretical analysis of the topology and use rationalizability as the solution concept, we focus on operationalizing BE within a computational framework. Furthermore, our solution concept is that of best response to one's beliefs.

## 3. Background

We briefly review interactive influence diagrams (I-ID) for two-agent interactions followed by their extension to dynamic settings, I-DIDs (Doshi et al., 2009). Both these formalisms allow modeling the other agent and to use that information in the decision making of the subject agent.

We illustrate the formalisms and our approaches in the context of the multiagent tiger problem (Gmytrasiewicz & Doshi, 2005) – a two-agent generalization of the well-known single agent tiger problem (Kaelbling et al., 1998). In this problem, two agents, $i$ and $j$, face two closed doors one of which hides a tiger while the other hides a pot of gold. An agent gets rewarded for opening the door that hides the gold but gets penalized for opening the door leading to the tiger. Each agent may open the left door (action denoted by OL), open the right door (OR), or listen (L). On listening, an agent may hear the tiger growling either from the left (observation denoted by GL) or from the right (GR). Additionally, the agent hears creaks emanating from the direction of the door that was possibly opened by the other agent – creak from the left (CL) or creak from right (CR) – or silence (S) if no door was opened. All observations are assumed to be noisy. If any door is opened by an agent, the tiger appears behind any of the two doors randomly in the next time step. While the actions of the other agent do not directly affect the reward for an agent, they may potentially change the location of the tiger. This formulation of the problem differs from that of Nair et al. (2003) in the presence of door creaks and that it is not cooperative.

### 3.1 Interactive Dynamic Influence Diagrams

Influence diagrams (Tatman & Shachter, 1990) typically contain chance nodes which represent the random variables modeling the physical state, $S$, and the agent's observations, $O_i$, among other aspects of the problem; decision nodes that model the agent's actions, $A_i$; and utility nodes that model the agent's reward function, $R_i$. In addition to these nodes, I-IDs for an agent $i$ include a new type of node called the *model node*. This is the hexagonal node, $M_{j,l-1}$, in Fig. 2, where $j$ denotes the other agent and $l-1$ is the *strategy level*, which allows for a nested modeling of $i$ by the other agent $j$. Agent $j$'s level is one less than that of $i$, which is consistent with previous hierarchical modeling in game theory (Aumann, 1999a; Brandenburger & Dekel, 1993) and decision theory (Gmytrasiewicz & Doshi, 2005). Additionally, a level 0 model is an ID or a flat probability distribution. We note that the probability distribution over the chance node, $S$, and the model node together represents agent $i$'s belief over its *interactive state space*. In addition to the model node, I-IDs differ from IDs by having a chance node, $A_j$, that represents the distribution over the other agent's actions, and a dashed link, called a *policy link*.

The model node contains as its values the alternative computational models ascribed by $i$ to the other agent. The policy link denotes that the distribution over $A_j$ is contingent on the models in the model node. We denote the set of these models by $\mathcal{M}_{j,l-1}$, and an individual model of $j$ as, $m_{j,l-1} = \langle b_{j,l-1}, \hat{\theta}_j \rangle$, where $b_{j,l-1}$ is the level $l-1$ belief, and $\hat{\theta}_j$ is the agent's *frame* encompassing the decision, observation and utility nodes. A model in the model node may itself be an I-ID or ID, and the recursion terminates when a model is an ID or a flat probability distribution over the actions.

We observe that the model node and the dashed policy link that connects it to the chance node, $A_j$, could be represented as shown in Fig. 3($a$) leading to a flat ID shown in Fig. 3($b$). The decision node of each level $l-1$ I-ID is transformed into a chance node. Specifically, if $\mathsf{OPT}(m_{j,l-1}^1)$ is the set of optimal actions obtained by solving the I-ID (or ID) denoted by $m_{j,l-1}^1$, then $Pr(a_j \in$ $A_j^1) = \frac{1}{|\mathsf{OPT}(m_{j,l-1}^1)|}$ if $a_j \in \mathsf{OPT}(m_{j,l-1}^1)$, 0 otherwise. The different chance nodes ($A_j^1$, $A_j^2$) – one for each model – and additionally, the chance node labeled $Mod[M_j]$ form the parents of the chance node, $A_j$. There are as many action nodes as the number of models in the support of agent $i$'s beliefs. The conditional probability table (CPT) of the chance node, $A_j$, is a multiplexer that assumes the distribution of each of the action nodes ($A_j^1$, $A_j^2$) depending on the value of $Mod[M_j]$. In other words, when $Mod[M_j]$ has the value $m_{j,l-1}^1$, the chance node $A_j$ assumes the distribution of the node $A_j^1$, and $A_j$ assumes the distribution of $A_j^2$ when $Mod[M_j]$ has the value $m_{j,l-1}^2$. The distribution over $Mod[M_j]$ is $i$'s belief over $j$'s models given the state.

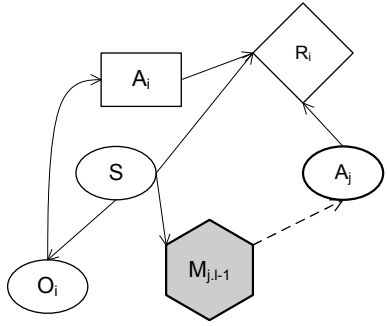

Figure 2: A generic level $l > 0$ I-ID for agent $i$ situated with one other agent $j$. The shaded hexagon is the model node ($M_{j,l-1}$) and the dashed arrow is the policy link.

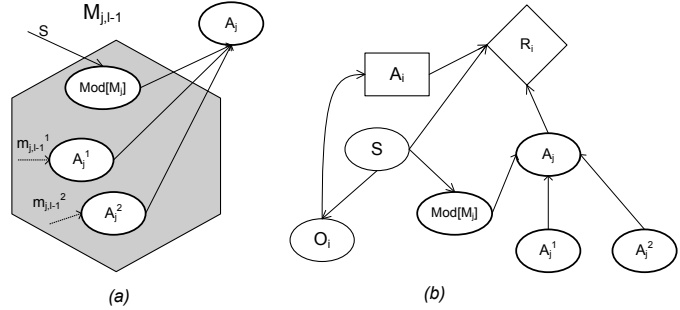

Figure 3: ($a$) Representing the model node and policy link using chance nodes and dependencies between them. The decision nodes of the lower-level I-IDs or IDs ($m_{j,l-1}^1$, $m_{j,l-1}^2$; superscript numbers serve to distinguish the models) are mapped to the corresponding chance nodes ($A_j^1$, $A_j^2$) respectively, which is indicated by the dotted arrows. Depending on the value of node, $Mod[M_j]$, distribution of each of the chance nodes is assigned to node $A_j$ with some probability. ($b$) The transformed flat ID with the model node and policy link replaced as in ($a$).

For more than two agents, we add a model node and a chance node representing the distribution over an agent's action linked together using a policy link, for each other agent. Interactions among others such as coordination or team work could be considered by utilizing models, which predict the *joint* behavior of others, in a distinct model node and possibly updating such models. For example, the joint behavioral models could be graphical analogs of decentralized POMDPs. Other settings involving agents acting independently or some of them being cooperative while others adversarial may be represented as well, and is a topic of research under study. As an aside, Doshi et al. (2009) show how I-IDs relate to NIDs (Gal & Pfeffer, 2008).

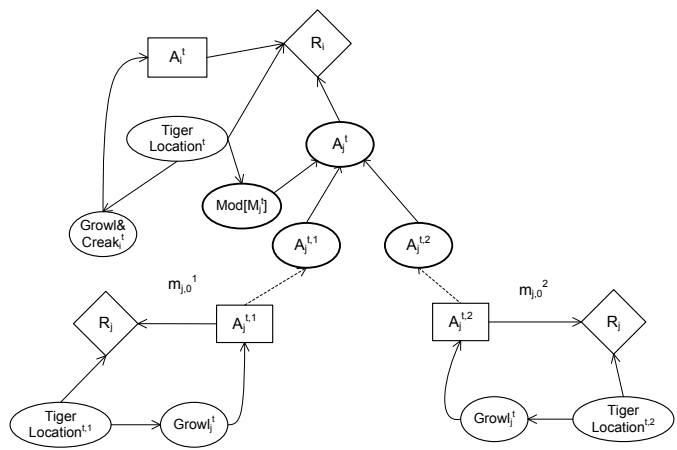

Figure 4: Level 1 I-ID of $i$ for the multiagent tiger problem. Solutions of two level 0 models (IDs) of $j$ map to the chance nodes, $A_j^{t,1}$ and $A_j^{t,2}$, respectively (illustrated using dotted arrows), transforming the I-ID into a flat ID. The two models differ in the distribution over the chance node, TigerLocation$^t$.

We setup the I-ID for the multiagent tiger problem described previously, in Fig. 4. We discuss the CPTs of the various nodes in Appendix B.1. While the I-ID contains two models of $j$, there would be as many action nodes of $j$ if there were more models.

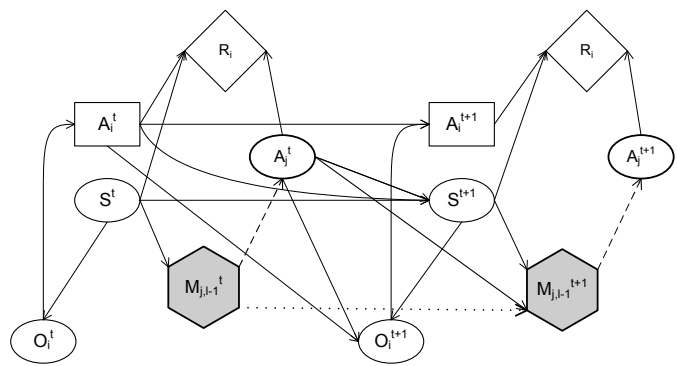

Figure 5: A generic two time-slice level $l$ I-DID for agent $i$. Notice the dotted model update link that denotes the update of the models of $j$ and of the distribution over the models, over time.

I-DIDs extend I-IDs to allow sequential decision making over multiple time steps. We depict a general, two time-slice I-DID in Fig. 5. In addition to the model nodes and the dashed policy link, what differentiates an I-DID from a DID is the *model update link* shown as a dotted arrow in Fig. 5. We briefly explain the semantics of the model update next.

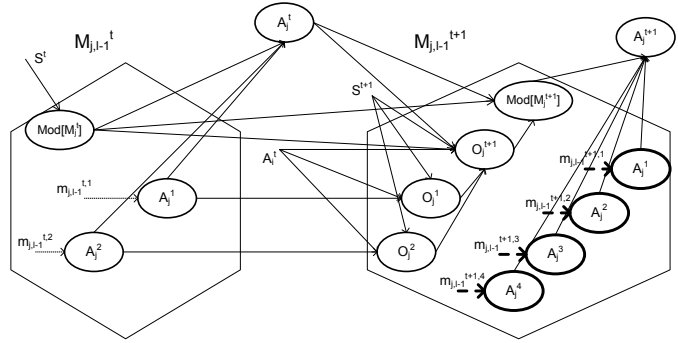

Figure 6: The semantics of the model update link. Notice the growth in the number of models in the model node at $t+1$ shown in bold (superscript numbers distinguish the different models). Models at $t+1$ reflect the updated beliefs of $j$ and their solutions provide the probability distributions for the action nodes.

Agents in a multiagent setting may act and make observations, which changes their beliefs. Therefore, the update of the model node over time involves two steps: First, given the models at time $t$, we identify the updated set of models that reside in the model node at time $t+1$. Because the agents act and receive observations, their models are updated to reflect their changed beliefs. Since the set of optimal actions for a model could include all the actions, and the agent may receive any one of $|\Omega_j|$ possible observations where $\Omega_j$ is the set of $j$'s observations, the updated set at time step $t+1$ will have up to $|\mathcal{M}_{j,l-1}^t||A_j||\Omega_j|$ models. Here, $|\mathcal{M}_{j,l-1}^t|$ is the number of models at time step $t$, $|A_j|$ and $|\Omega_j|$ are the largest spaces of actions and observations respectively, among all the models. The CPT of chance node $Mod[M_{j,l-1}^{t+1}]$ encodes the indicator function, $\tau(b_{j,l-1}^t, a_j^t, o_j^{t+1}, b_{j,l-1}^{t+1})$, which is 1 if the belief $b_{j,l-1}^t$ in a model $m_{j,l-1}^t$ using the action $a_j^t$ and observation $o_j^{t+1}$ updates to $b_{j,l-1}^{t+1}$ in a model $m_{j,l-1}^{t+1}$; otherwise it is 0. Second, we compute the new distribution over the updated models given the original distribution and the probability of the agent performing the action and receiving the observation that led to the updated model. The dotted model update link in the I-DID may be implemented using standard dependency links and chance nodes, as shown in Fig. 6 transforming the I-DID into a flat DID.

In Fig. 7, we show the two time-slice flat DID with the model nodes and the model update link replaced by the chance nodes and the relationships between them. Chance nodes and dependency links not in bold are standard, usually found in single agent DIDs.

Continuing with our illustration, we show the two time-slice I-DID for the multiagent tiger problem in Fig. 8. The model update link not only updates the number of $j$'s candidate models due to its action and observations of growl, it also updates the probability distribution over these models. The model update link in the I-DID is implemented using standard dependency links as shown in Fig. 9. For the sake of clarity, we illustrate the update of a single model of $j$ contained in the model node at time $t$.

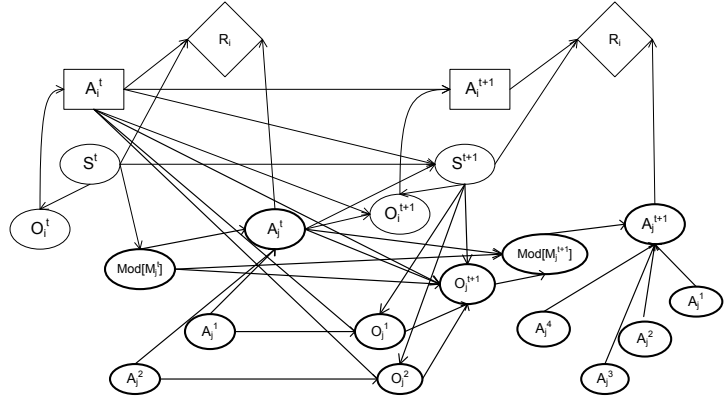

Figure 7: A flat DID obtained by replacing the model nodes and model update link in the I-DID of Fig. 5 with the chance nodes and the relationships (in bold) as shown in Fig. 6. The lower-level models are solved to obtain the distributions for the chance action nodes.

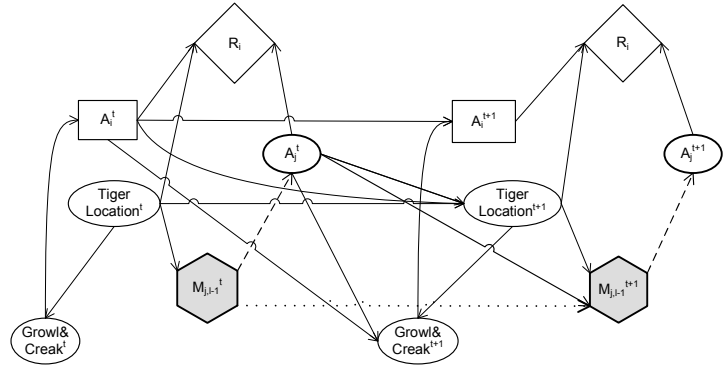

Figure 8: Two time-slice level $l$ I-DID of $i$ for the multiagent tiger problem. Shaded model nodes contain the different models of $j$.

## 3.2 Behavioral Equivalence and Model Solution

Although the space of possible models is very large, not all models need to be considered by agent $i$ in the model node. As we mentioned previously, models that are *BE* (Rathnasabapathy et al., 2006; Pynadath & Marsella, 2007) could be pruned and a single representative model considered. This is because the solution of the subject agent's I-DID is affected by the predicted behavior of the other agent; thus we need not distinguish between behaviorally equivalent models. We define BE more formally below:

**Definition 1** (Behavioral equivalence). *Two models, $m_{j,l-1}$ and $m'_{j,l-1}$, of the other agent, $j$, are behaviorally equivalent if, $\mathsf{OPT}(m_{j,l-1}) = \mathsf{OPT}(m'_{j,l-1})$, where $\mathsf{OPT}(\cdot)$ denotes the solution of the model that forms the argument.*

Thus, BE models are those whose behavioral predictions for the agent are identical.

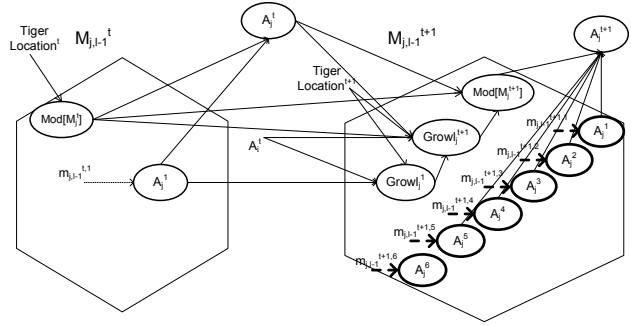

Figure 9: Because agent $j$ in the tiger problem may receive any one of six possible observations given the action prescribed by its model, a single model in the model node at time $t$ could lead to six distinct models at time $t + 1$.

The solution of an I-DID (and I-ID) is implemented recursively down the levels as shown in Fig. 10. In order to solve a level 1 I-DID of horizon $T$, we start by solving the level 0 models, which may be traditional DIDs of horizon $T$. Their solutions provide probability distributions over the other agents' actions, which are entered in the corresponding action nodes found in the model node of the level 1 I-DID at the corresponding time step (lines 3-5). Subsequently, the set of $j$'s models is minimized by excluding the BE models (line 6).

The solution method uses the standard look-ahead technique, projecting the agent's action and observation sequences forward from the current belief state, and finding the possible beliefs that $i$ could have in the next time step (Russell & Norvig, 2010). Because agent $i$ has a belief over $j$'s models as well, the look-ahead includes finding out the possible models that $j$ could have in the future. Consequently, each of $j$'s level 0 models represented using a standard DID must be solved in the first time step up to horizon $T$ to obtain its optimal set of actions. These actions are combined with the set of possible observations that $j$ could make in that model, resulting in an updated set of candidate models (that include the updated beliefs) that could describe the behavior of $j$. $SE(b_j^t, a_j, o_j)$ is an abbreviation for the belief update (lines 8-13). Beliefs over these updated set of candidate models are calculated using the standard inference methods through the dependency links between the model nodes shown in Fig. 6 (lines 15-18). Agent $i$'s I-DID is expanded across all time steps in this manner. We point out that the algorithm in Fig. 10 may be realized with the help of standard implementations of DIDs such as HUGIN EXPERT (Andersen & Jensen, 1989). The solution is a policy tree that prescribes the optimal action(s) to perform for agent $i$ initially given its belief, and the actions thereafter conditional on its observations.

## 4. Discriminative Model Updates

Solving I-DIDs is computationally intractable due to not only the large space and complexity of models ascribed to $j$, but also due to the exponential growth in candidate models of $j$ over time. This growth leads to a disproportionate increase in the interactive state space over time. We begin by introducing a set of models that is *minimal* in a sense and describe a method for generating this set. The minimal set is analogous to one of the notions of a minimal mental model space as described by

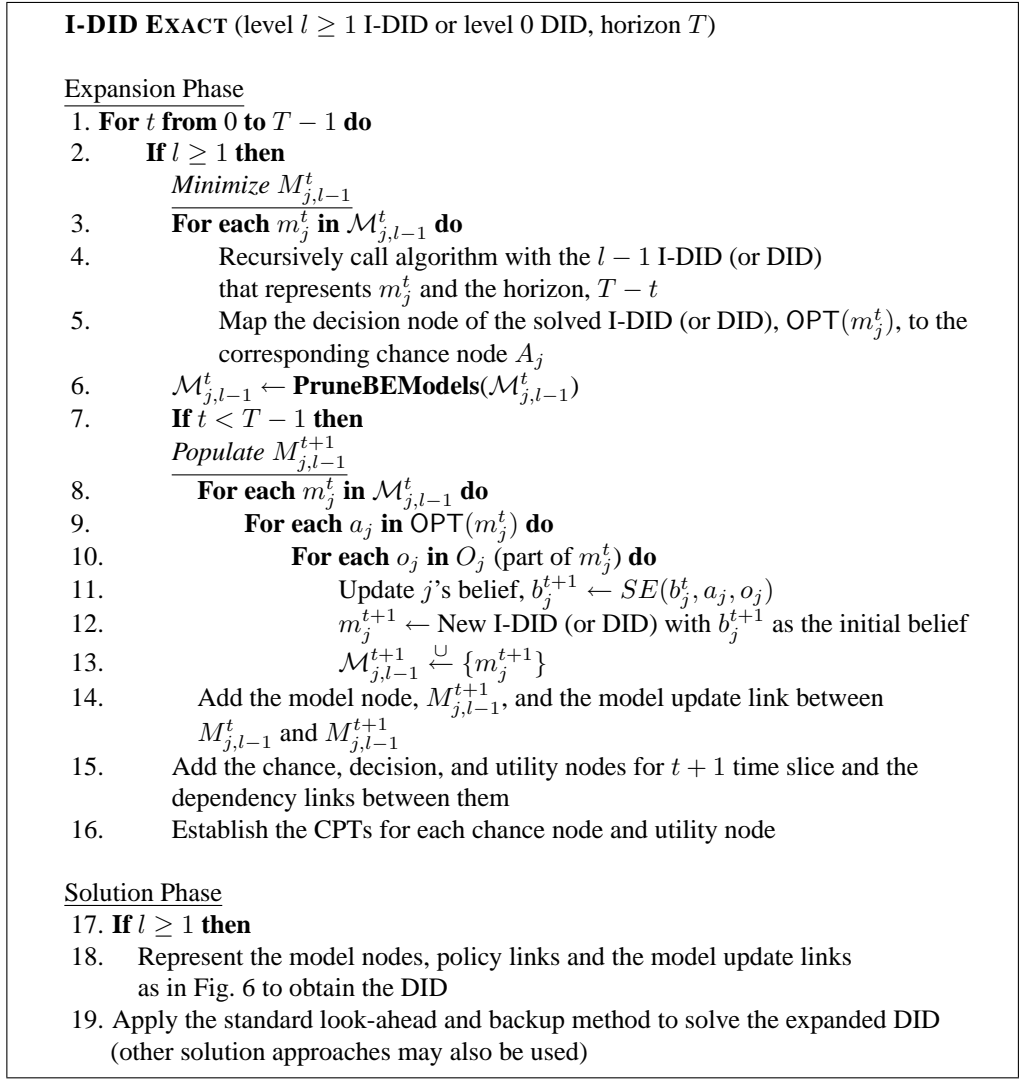

Figure 10: Algorithm for exactly solving a level $l \geq 1$ I-DID or level 0 DID expanded over $T$ time steps.

Pynadath and Marsella (2007). We assume that models of the other agent differ only in their beliefs and that the other agent's frame is known. We later discuss in Section 7 the impact of the frame being unknown as well. For clarity, we continue to focus on two-agent interactions, and discuss extensions of the techniques presented here in Section 7 as well.

## 4.1 Behaviorally Minimal Model Set

Given the set of models, $\mathcal{M}_{j,l-1}$, of the other agent, $j$, in a model node we define a corresponding *behaviorally minimal* set of models:

**Definition 2** (Behaviorally minimal set). *Define a minimal set of models, $\hat{\mathcal{M}}_{j,l-1}$, as the largest subset of $\mathcal{M}_{j,l-1}$, such that for each model, $m_{j,l-1} \in \hat{\mathcal{M}}_{j,l-1}$, there exists no other model in $\hat{\mathcal{M}}_{j,l-1}$ that is BE to $m_{j,l-1}$.*

Here, BE is as defined in Def. 1. We say that $\hat{\mathcal{M}}_{j,l-1}$ (behaviorally) *minimizes* $\mathcal{M}_{j,l-1}$. As we illustrate in Fig. 11 using the tiger problem (Kaelbling et al., 1998), the set $\hat{\mathcal{M}}_{j,l-1}$ that minimizes $\mathcal{M}_{j,l-1}$ comprises of all the behaviorally *distinct* representatives of the models in $\mathcal{M}_{j,l-1}$ and only these models. Because any model from a group of BE models may be selected as the representative in $\hat{\mathcal{M}}_{j,l-1}$, a minimal set corresponding to $\mathcal{M}_{j,l-1}$ is not unique, although its cardinality remains fixed.

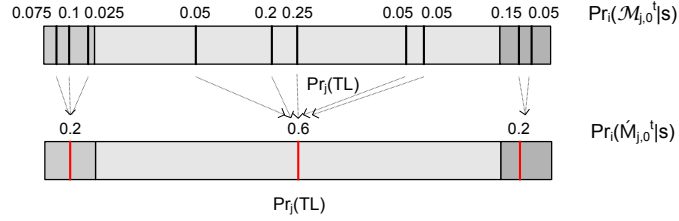

Figure 11: Illustration of a minimal set using the tiger problem. Black vertical lines denote the beliefs contained in different models of agent $j$ included in model node, $M_{j,0}$. Decimals on top indicate $i$'s probability distribution over $j$'s models, $Pr_i(\mathcal{M}_{j,0}^t|s)$. In order to form a behaviorally minimal set, $\hat{\mathcal{M}}_{j,0}^t$, we select a representative model from each BE group of models (models in differently shaded regions). Agent $i$'s distribution over the models in $\hat{\mathcal{M}}_{j,0}^t$ is obtained by summing the probability mass assigned to the individual models in each region. Note that $\hat{\mathcal{M}}_{j,0}^t$ is not unique because any one model within a shaded region could be selected for inclusion in it.

Agent $i$'s probability distribution over the minimal set, $\hat{\mathcal{M}}_{j,l-1}$, conditioned on the physical state is obtained by summing the probability mass over BE models in $\mathcal{M}_{j,l-1}$ and assigning the accumulated probability to the representative model in $\hat{\mathcal{M}}_{j,l-1}$. Formally, let $\hat{m}_{j,l-1} \in \hat{\mathcal{M}}_{j,l-1}$, then:

$$\hat{b}_i(\hat{m}_{j,l-1}|s) = \sum_{m_{j,l-1} \in \mathbb{M}_{j,l-1}} b_i(m_{j,l-1}|s) \qquad (1)$$

where $\mathbb{M}_{j,l-1} \subseteq \mathcal{M}_{j,l-1}$ is the set of BE models to which the representative $\hat{m}_{j,l-1}$ belongs. Thus, if $\hat{\mathcal{M}}_{j,l-1}$ minimizes $\mathcal{M}_{j,l-1}$, then Eq. 1 shows how we may obtain the probability distribution over $\hat{\mathcal{M}}_{j,l-1}$ at some time step, given $i$'s belief distribution over models in the model node at that step (see Fig. 11).

The behaviorally minimal set together with the probability distribution over it has an important property: Solution of an I-DID remains unchanged when the models in a model node and the distribution over the models are replaced by the corresponding minimal set and the distribution over it, respectively. In other words, transforming the set of models in the model node into its minimal set preserves the solution. Proposition 1 states this formally:

**Proposition 1.** *Let $X : \Delta(\mathcal{M}_{j,l-1}) \rightarrow \Delta(\hat{\mathcal{M}}_{j,l-1})$ be a mapping defined by Eq. 1, where $\mathcal{M}_{j,l-1}$ is the space of models in a model node and $\hat{\mathcal{M}}_{j,l-1}$ minimizes it. Then, applying $X$ preserves the solution of the I-DID.*

Proof of Proposition 1 is given in Appendix A. Proposition 1 allows us to show that $\hat{\mathcal{M}}_{j,l-1}$ is indeed minimal given $\mathcal{M}_{j,l-1}$ with respect to the solution of the I-DID.

**Corollary 1.** *$\hat{\mathcal{M}}_{j,l-1}$ in conjunction with $X$ is a sufficient solution-preserving subset of models found in $\mathcal{M}_{j,l-1}$.*

Proof of this corollary follows directly from Proposition 1. Notice that the subset continues to be solution preserving when we additionally augment $\hat{\mathcal{M}}_{j,l-1}$ with models from $\mathcal{M}_{j,l-1}$.

As the number of models in the minimal set is, of course, no more than in the original set and typically much less, solution of the I-DID is often computationally much less intensive when the model set is replaced with its behaviorally minimal counterpart.

### 4.2 Discrimination Using Policy Graphs

A straightforward way of obtaining $\hat{\mathcal{M}}_{j,l-1}$ *exactly* at any time step is to first ascertain the BE groups of models. This requires us to solve the I-DIDs or DIDs representing the models, then select a representative model from each BE group to include in $\hat{\mathcal{M}}_{j,l-1}$, and prune all others which have the same solution as the representative.

#### 4.2.1 APPROACH

Given the set of $j$'s models, $\mathcal{M}_{j,l-1}$, at time $t(=0)$, we present a technique for generating the minimal sets at subsequent time steps in the I-DID. *We first observe that behaviorally distinct models at time $t$ may result in updated models at $t + 1$ that are BE.* Hence, our approach is to select at time step $t$ only those models for updating which will result in predictive behaviors that are distinct from others in the updated model space at $t + 1$. Models that will result in predictions on update which are identical to those of other existing models at $t + 1$ are not selected for updating. Consequently, the resulting model set at $t + 1$ is minimal.

We do this by solving the individual I-DIDs or DIDs in $\mathcal{M}_{j,l-1}^t$. Solutions to DIDs or I-DIDs are policy trees, which may be merged bottom up to obtain a *policy graph*, as we demonstrate in Fig. 12. Seuken and Zilberstein (2007) reuse subtrees of smaller horizon by linking to them using pointers while forming policy trees for the next horizon in the solution of decentralized POMDPs. The net effect is the formation of a policy graph similar to ours thereby providing an alternative to our approach of solving the individual models to first obtain the complete policy trees and then merge post hoc. We adopt the latter approach because the individual models, which are DIDs, when solved using available implementations produce complete policy trees. The following proposition gives the complexity of merging the policy trees to obtain the policy graph.

**Proposition 2** (Complexity of tree merge)**.** *The worst-case complexity of the procedure for merging policy trees to form a policy graph is $\mathcal{O}((|\Omega_j|^{T-1})^2|\mathcal{M}_j|^2)$, where $T$ is the horizon.*

*Proof.* The complexity of the policy tree merge procedure is proportional to the number of comparisons that are made between parts of policy trees to ascertain their similarity. Because the procedure follows a bottom-up approach and the leaf level has the largest number of nodes, the maximum number of comparisons are made between leaf nodes. The worst case occurs when none of the leaf nodes of the different policy trees can be merged. Note that this precludes the merger of upper parts of the policy trees as well. Each policy tree may contain up to $|\Omega_j|^{T-1}$ leaf nodes, where $T$ is the horizon. Hence, at most $\mathcal{O}((|\Omega_j|^{T-1})^2|\mathcal{M}_j|^2)$ comparisons are made, where $\mathcal{O}(|\mathcal{M}_j|^2)$ is the number of pairs in the model set. [1] The case when none of the leaf nodes merge must occur when the models are behaviorally distinct, and they form a minimal set, $\hat{\mathcal{M}}_j$. In other words, $\mathcal{M}_j = \hat{\mathcal{M}}_j$. ∎

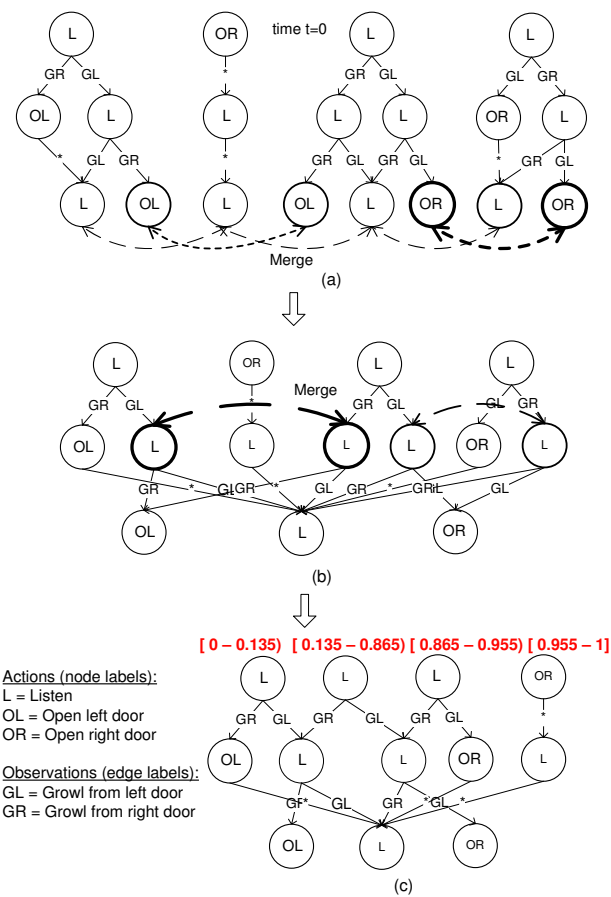

Figure 12: $(a)$ Example policy trees obtained by solving four models of $j$ for the tiger problem. Beginning bottom up, we may merge the four L nodes, two OR nodes and two OL nodes respectively to obtain the graph in $(b)$. Because the two policy trees of two steps rooted at L (bold circle) are identical, and so are the two policy trees rooted at L (rightmost), we may merge them, respectively, to obtain the policy graph in $(c)$. Nodes at $t = 0$ are annotated with ranges of $Pr_j(TL)$.

---

1. If we assume an ordering of the observations (edge labels) thereby ordering the tree, two policy trees may be sufficiently compared in $\mathcal{O}(|\Omega_j|^{T-1})$ time.

Each node in the policy graph represents an action to be performed by the agent and edges represent the agent's observations. As is common with policy graphs in POMDPs, we associate with each node at time $t = 0$, a range of beliefs for which the corresponding action is optimal (see Fig. 12$(c)$). This range may be obtained by computing the value of executing the policy tree rooted at each node at $t = 0$ in the graph and starting from each physical state. This results in a vector of values for each policy tree, typically called the $\alpha$-vector. Intersecting the $\alpha$-vectors and projecting the intersections on the belief simplex provides us with the boundaries of the needed belief ranges.

We utilize the policy graph to discriminate between model updates. For clarity, we formally define a policy graph next.

**Definition 3** (Policy graph). *Define a policy graph as:*

$$PG = \langle \mathcal{V}, A, \mathcal{E}, \Omega, \mathcal{L}_v, \mathcal{L}_e \rangle$$

*where $\mathcal{V}$ is the set of vertices (nodes); $A$ is the set of actions which form the node labels; $\mathcal{E}$ is the set of ordered pairs of vertices (edges); $\Omega$ is the set of observations which form the edge labels; $\mathcal{L}_v : \mathcal{V} \to A$ assigns to each vertex an action from the set of actions, $A$ (node label); and $\mathcal{L}_e : \mathcal{E} \to \Omega$ assigns to each edge an observation from the set of observations, $\Omega$ (edge label). $\mathcal{L}_e$ follows the property that no two edges whose first elements are identical (begin at the same vertex) are assigned the same observation.*

Notice that a policy graph augments a regular graph with meaningful node and edge labels. For a policy graph, $PG$, we also define the transition function, $\mathcal{T}_p : \mathcal{V} \times \Omega \to \mathcal{V}$, for convenience. $\mathcal{T}_p(v, o)$ returns the vertex, $v'$, such that $\{v, v'\} \in \mathcal{E}$ and $\mathcal{L}_e(\{v, v'\}) = o$.

*Our insight is that $\mathcal{T}_p(v, o)$ is the root node of a policy tree that represents the predictive behavior for the model updated using the action $\mathcal{L}_v(v)$ and observation $o$.* As we iterate over $j$'s models in the model node at time $t$ in the expansion phase while solving the I-DID, we utilize $\mathcal{T}_p$ in deciding whether to update a model.

We first combine the policy trees obtained by solving the models in node $M_{j,l-1}^t$ to obtain the policy graph, $PG$, as shown in Fig. 12. Let $v$ be the vertex in $PG$ whose action label, $\mathcal{L}_v(v)$, represents the rational action for $m_{j,l-1} \in \mathcal{M}_{j,l-1}^t$. We can ascertain this by simply checking whether the belief in $m_{j,l-1}$ falls within the belief range associated with a node. For every observation $o \in \mathcal{L}_e(\{v, \cdot\})$, we update the model, $m_{j,l-1}$, using action $\mathcal{L}_v(v)$ and observation $o$, if $v' = \mathcal{T}_p(v, o)$ has not been encountered previously for this or any other model. We illustrate this below:

**Example 1** (Model update). *Consider the level 0 models of $j$ in the model node at time $t$, $\mathcal{M}_{j,0}^t = \{\langle 0.01, \hat{\theta}_j \rangle, \langle 0.5, \hat{\theta}_j \rangle, \langle 0.05, \hat{\theta}_j \rangle\}$, for the multiagent tiger problem. Recall that in a model of $j$, such as $\langle 0.01, \hat{\theta}_j \rangle$, 0.01 is $j$'s belief (over TL) and $\hat{\theta}_j$ is its frame. From the PG in Fig. 12(c), the leftmost node prescribing the action $L$ is optimal for the first and third models, while the second node also prescribing $L$ is optimal for the second model. Beginning with model, $\langle 0.01, \hat{\theta}_j \rangle$, $\mathcal{T}_p(v, GL) = v_1$ (where $\mathcal{L}_v(v_1) = L$) and $\mathcal{T}_p(v, GR) = v_2$ ($\mathcal{L}_v(v_2) = OL$). Since this is the first model we consider, it will be updated using $L$ and both observations resulting in two models in $\mathcal{M}_{j,0}^{t+1}$. For the model, $\langle 0.5, \hat{\theta}_j \rangle$, if $v'$ is the optimal node ($\mathcal{L}_v(v') = L$), $\mathcal{T}_p(v', GR) = v_1$, which has been encountered previously. Hence, the model will not be updated using $L$ and $GR$, although it will be updated using $L$ and $GL$.*

Intuitively, for a model, $m_{j,l-1}$, if node $v_1 = \mathcal{T}_p(v, o)$ has been obtained previously for this or some other model and action-observation combination, then the update of $m_{j,l-1}$ will be BE to the previously updated model (both will have the same policy tree rooted at $v_1$). Hence, $m_{j,l-1}$ need not be updated using the observation $o$. *Because we do not permit updates that will lead to BE models, the set of models obtained at $t + 1$ is minimal.* Applying this process analogously to models in the following time steps will lead to minimal sets at all subsequent steps and nesting levels.

### 4.2.2 Approximation

We may gain further efficiency by avoiding the solution of all models in the model node at the first time step. One way of doing this is to randomly select $K$ models of $j$, such that $K \ll |\mathcal{M}_{j,l-1}^0|$. Solution of the models will result in $K$ policy trees, which could be combined as shown in Fig. 12 to form a policy graph. This policy graph is utilized to discriminate between the model updates. Notice that the approach becomes exact if the optimal solution of each model in $\mathcal{M}_{j,l-1}^0$ is identical to that of one of the $K$ models. Because the $K$ models are selected randomly, this assumption is implausible and the approach is likely to result in a substantial loss of optimality that is mediated by $K$.

We propose a simple but effective refinement that mitigates the loss. Recall that models whose beliefs are spatially close are likely to be BE (Rathnasabapathy et al., 2006). Each of the remaining $|\mathcal{M}_{j,l-1}^0| - K$ models whose belief is not within $\epsilon \geq 0$ of the belief of any of the $K$ models will also be solved. This additional step makes it more likely that all the behaviorally distinct solutions will be generated and included in forming the policy graph. If $\epsilon = 0$, all models in the model node will be solved leading to the exact solution, while increasing $\epsilon$ reduces the number of solved models beyond $K$. One measure of distance between belief points is the $L_1$ based metric, though other metrics such as the Euclidean distance may also be used.

### 4.3 Transfer of Probability Mass

Notice that a consequence of not updating models using some action-observation combination is that the probability mass that would have been assigned to the updated model in the model node at $t+1$ is lost. Disregarding this probability mass may introduce error in the optimality of the solution.

We did not perform the update because a model that is BE to the potentially updated model already exists in the model node at time $t+1$. We could avoid the error by transfering the probability mass that would have been assigned to the updated model on to the BE model.

As we mentioned previously, the node $Mod[M_{j,l-1}^{t+1}]$ in the model node $M_{j,l-1}^{t+1}$, has as its values the different models ascribed to agent $j$ at time $t + 1$. The CPT of $Mod[M_{j,l-1}^{t+1}]$ implements the function $\tau(b_{j,l-1}^t, a_j^t, o_j^{t+1}, b_{j,l-1}^{t+1})$, which is 1 if $b_{j,l-1}^t$ in the model $m_{j,l-1}^t$ updates to $b_{j,l-1}^{t+1}$ in model $m_{j,l-1}^{t+1}$ using the action-observation combination, otherwise it is 0. Let $m_{j,l-1}^{t+1'} = \langle b_{j,l-1}^{t+1'}, \hat{\theta}_j \rangle$ be the model that is BE to $m_{j,l-1}^{t+1}$. In order to transfer the probability mass to this model if the update is pruned, we modify the CPT of $Mod[M_{j,l-1}^{t+1}]$ to indicate that $m_{j,l-1}^{t+1'}$ is the model that results from updating $b_{j,l-1}^t$ with action, $a_j^t$ and observation $o_j^{t+1}$. This has the desired effect of transfering the probability that would have been assigned to the updated model (Fig. 6) on to $m_{j,l-1}^{t+1'}$ in the model node at time $t + 1$.

### 4.4 Algorithm

We present the discriminative update based algorithm for solving a level $l \geq 1$ I-DID (as well as a level 0 DID) in Fig. 13. The algorithm differs from the exact approach (Fig. 10) in the expansion phase. In addition to a two time-slice level $l$ I-DID and horizon $T$, the algorithm takes as input the number of random models to be solved initially, $K$, and the distance, $\epsilon$. Following Section 4.2, we begin by randomly selecting $K$ models to solve (lines 2-5). For each of the remaining models, we identify one of the $K$ solved model whose belief is spatially the closest (ties broken randomly). If the proximity is within $\epsilon$, the model is not solved – instead, the previously computed solution is assigned to the corresponding action node of the model in the model node, $M_{j,l-1}^0$ (lines 6-12). Subsequently, all models in the model node are associated with their respective solutions (policy trees), which are merged to obtain the policy graph (line 13), as illustrated in Fig. 12.

In order to populate the model node of the next time step, we identify the node $v$ in $PG$ that represents the optimal action for a model at time $t$. The model is updated using the optimal action $a_j$ ($= \mathcal{L}_v(v)$) and each observation $o_j$ only if the node, $v' = \mathcal{T}_p(v, o_j)$ has not been encountered in previous updates (lines 16-23). Given a policy graph, evaluating $\mathcal{T}_p(v, o_j)$ is a constant time operation. Otherwise, as mentioned in Section 4.3, we modify the CPT of node, $Mod[M_{j,l-1}^{t+1}]$, to transfer the probability mass to a BE model (line 25). Consequently, model nodes at subsequent time steps in the expanded I-DID are likely populated with minimal sets. Given the expanded I-DID, its solution may proceed in a straightforward manner as shown in Fig. 10.

### 4.5 Computational Savings and Prediction Error Bound

The primary complexity of solving I-DIDs is due to the large number of models that must be solved over $T$ time steps. At some time step $t$, there could be $|\mathcal{M}_{j,l-1}^0|(|A_j||\Omega_j|)^t$ many models of the other agent $j$, where $|\mathcal{M}_{j,l-1}^0|$ is the number of models considered initially. The nested modeling further contributes to the complexity since solutions of each model at level $l-1$ requires solving the lower level $l-2$ models, and so on recursively up to level 0. In an $N{+}1$ agent setting, if the number of models considered at each level for an agent is bound by $|\mathcal{M}|$, then solving an I-DID at level $l$ requires the solutions of $\mathcal{O}((N|\mathcal{M}|)^l)$ many models. Discriminating between model updates reduces the number of agent models at each level to at most the size of the behaviorally minimal set, $|\hat{\mathcal{M}}^t|$, while incurring the worst-case complexity of $\mathcal{O}((|\Omega|^{T-1})^2|\mathcal{M}|^2)$ in forming the policy graph (Proposition 2). Consequently, we need to solve at most $\mathcal{O}((N|\hat{\mathcal{M}}^*|)^l)$ number of models at each non-initial time step, where $\hat{\mathcal{M}}^*$ is the largest of the minimal sets across levels. [2] This is in comparison to $\mathcal{O}((N|\mathcal{M}|)^l)$, where $\mathcal{M}$ grows exponentially over time. In general, $\hat{\mathcal{M}} \ll \mathcal{M}$, resulting in a substantial reduction in the computation. Additionally, a reduction in the number of models in the model node also reduces the size of the interactive state space, which makes solving the I-DID more efficient.

If we choose to solve all models in the initial model node, $M_{j,l-1}^0$, in order to form the policy graph, all sets of models at subsequent time steps will indeed be minimal. Consequently, there is no loss in the optimality of the solution of agent $i$'s level $l$ I-DID.

For the case where we select $K < |\mathcal{M}_{j,l-1}^0|$ models to solve, if $\epsilon$ is infinitesimally small, we will eventually solve all models resulting in no error. With increasing values of $\epsilon$, larger numbers

---

2. As we discuss in Section 7, we may group BE models across agents as well due to which the number of models to be solved further reduces.

---

**I-DID Approx BE** (level $l \geq 1$ I-DID or level 0 DID, $T$, $K$, $\epsilon$)

1. **If** $l \geq 1$ **then**

    *Selectively solve* $\mathcal{M}_{j,l-1}^0$

2.   Randomly select $K$ models from $\mathcal{M}_{j,l-1}^0$

3. **For each** $m_j^k$ **in the** $K$ models **do**

4.     Recursively call algorithm with the $l-1$ I-DID (or DID) that represents $m_j^k$,
    the horizon $T$, $K$, and $\epsilon$

5.     Map the decision node of the solved I-DID (or DID), $\mathsf{OPT}(m_j^k)$, to the chance node $A_j^k$

6. **For each** $m_j^{\bar{k}}$ **in the** $|\mathcal{M}_{j,l-1}^0| - K$ models **do**

7.     Find model among $K$ whose belief, $b_j^k$, is closest to $b_j^{\bar{k}}$ in $m_j^{\bar{k}}$

8.     **If** $||b_j^k - b_j^{\bar{k}}||_1 \leq \epsilon$ **then**

9.       Map the decision node of $\mathsf{OPT}(m_j^k)$ to the chance node, $A_j^{\bar{k}}$

10.     **else**

11.       Recursively call algorithm with the $l-1$ I-DID (or DID) that represents $m_j^{\bar{k}}$,
      the horizon, $T$, $K$, and $\epsilon$

12.       Map the decision node of the solved I-DID (or DID), $\mathsf{OPT}(m_j^{\bar{k}})$, to the chance node $A_j^{\bar{k}}$

13. Combine the solutions (policy trees) of all models bottom up to obtain the policy graph, $PG$

Expansion Phase

14. **For** $t$ **from** 0 **to** $T-2$ **do**

15.   **If** $l \geq 1$ **then**

    *Populate* $M_{j,l-1}^{t+1}$ *minimally*

16.     **For each** $m_j^t$ **in** $\mathcal{M}_{j,l-1}^t$ **do**

17.       **For each** $a_j$ **in** $\mathsf{OPT}(m_j^t)$ **do**

18.         **For each** $o_j$ **in** $\Omega_j$ (part of $m_j^t$) **do**

19.           $v \leftarrow$ vertex in $PG$ to which $m_j^t$ maps

20.           **If** $\mathcal{T}_p(v, o_j)$ not been encountered previously **then**

21.             Update $j$'s belief, $b_j^{t+1} \leftarrow SE(b_j^t, a_j, o_j)$

22.             $m_j^{t+1} \leftarrow$ New I-DID (or DID) with $b_j^{t+1}$ as belief

23.             $\mathcal{M}_{j,l-1}^{t+1} \overset{\cup}{\leftarrow} \{m_j^{t+1}\}$

24.           **else**

25.             Update CPT of $Mod[M_{j,l-1}^{t+1}]$ s. t. row $m_j^t$, $a_j$, $o_j$ has a 1 in column
            of BE model

26.     Add the model node, $M_{j,l-1}^{t+1}$, and the model update link between $M_{j,l-1}^t$ and $M_{j,l-1}^{t+1}$

27.   Add the chance, decision, and utility nodes for $t+1$ time slice and the dependency
  links between them

28.   Establish the CPTs for each chance node and utility node

The *solution phase* proceeds analogously as in Fig. 10

Figure 13: Algorithm for approximately solving a level $l \geq 1$ I-DID or level 0 DID expanded over $T$ time steps using discriminative model updates.

of models remain unsolved and could be erroneously associated with existing solutions. In the worst case, some of these models may be behaviorally distinct from all of the $K$ solved models. Therefore, the policy graph is a subgraph of the one in the exact case, and leads to sets of models that are subsets of the minimal sets. Additionally, lower-level models are solved approximately as well. While we seek to possibly bound the prediction error, it's impact on the optimality of agent $i$'s level $l$ I-DID is difficult to pinpoint. We formally define the error and discuss bounding it including some limitations on the bound, in Appendix A.

## 5. Grouping Models Using Action Equivalence

Grouping BE models may significantly reduce the given space of other agents' models in the model node without loss in optimality. We may further compact the space of models in the model node by observing that behaviorally distinct models may prescribe identical actions at a single time step. We may then group together these models into a single equivalence class. In comparison to BE, the equivalence class includes those models whose prescribed action for the *particular* time step is the same, and we call it *action equivalence*. We define it formally next.

### 5.1 Action Equivalence

Notice from Fig. 12($c$) that the policy graph contains multiple nodes labeled with the same action at time steps $t = 0$ and $t = 1$. The associated models while prescribing actions that are identical at a particular time step, differ in the entire behavior. We call these models *actionally equivalent*. For a general case, we define action equivalence (AE) below:

**Definition 4** (Action equivalence). *Two models, $m_{j,l-1}$ and $m'_{j,l-1}$, of the other agent are actionally equivalent at time step $t$ if $Pr(A_j^t) = Pr(A_j^{t'})$ where $Pr(a_j^t) = \frac{1}{|\mathsf{OPT}(m_{j,l-1})|}$ if $a_j^t \in \mathsf{OPT}(m_{j,l-1})$, 0 otherwise; and $Pr(a_j^{t'}) = \frac{1}{|\mathsf{OPT}(m'_{j,l-1})|}$ if $a_j^{t'} \in \mathsf{OPT}(m'_{j,l-1})$, 0 otherwise, as defined previously.*

Since AE may include behaviorally distinct models, it partitions the model space into fewer classes.

We show an example aggregation of AE models in Fig. 14. From the figure, the partition of the model set, $\mathcal{M}_{j,l-1}^t$, induced by AE at time step 0 is $\{\mathcal{M}_{j,l-1}^{t=0,1}, \mathcal{M}_{j,l-1}^{t=0,2}\}$, where $\mathcal{M}_{j,l-1}^{t=0,1}$ is the class of models in the model space whose prescribed action at $t = 0$ is $L$, and $\mathcal{M}_{j,l-1}^{t=0,2}$ is the class of models whose prescribed action at $t = 0$ is $OR$. Note that these classes include the BE models as well. Thus, all models in an AE class prescribe an identical action at that time step. Furthermore at $t = 1$, the partition consists of 3 AE classes and, at $t = 2$, the partition also consists of 3 singleton classes.

If $\mathcal{M}_{j,l-1}^{t,p}$ is an AE class comprising of models $m_{j,l-1}^t \in \mathcal{M}_{j,l-1}^{t,p}$, agent $i$'s conditional belief over it is obtained by summing over $i$'s conditional belief over its member models:

$$\hat{b}_i(\mathcal{M}_{j,l-1}^{t,p}|s) = \sum_{m_{j,l-1}^t \in \mathcal{M}_{j,l-1}^{t,p}} b_i(m_{j,l-1}^t|s) \tag{2}$$

### 5.2 Revised CPT of *Mod* Node and its Markov Blanket

Equation 2 changes the CPT of the node, $Mod[M_{j,l-1}^t]$, due to the aggregation. Chang and Fung (1991) note that a coarsening operation of this type will not affect the distributions that do not directly involve the model space if the joint probability distribution over the *Markov blanket* of node, $Mod[M_{j,l-1}^t]$, remains unchanged. In Fig. 15, we show the Markov blanket of $Mod[M_{j,l-1}^t]$. [3]

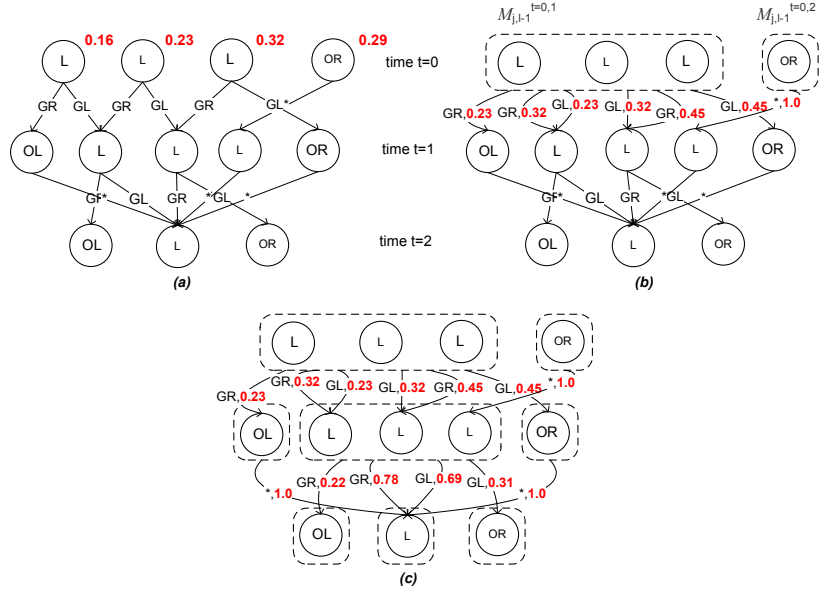

Figure 14: (*a*) Annotations are example probabilities for the models associated with the nodes. (*b*, *c*) We may group models that prescribe identical actions into classes as indicated by the dashed boxes. Probabilities on the edges represent the probability of transition from a class to a model given action and observation (ie., CPT of the next *Mod* node).

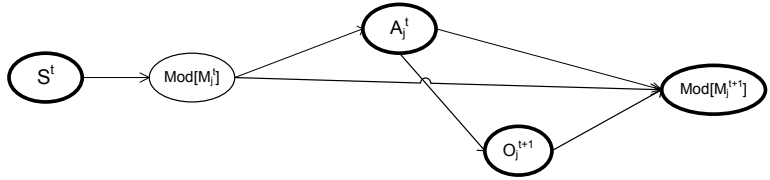

Figure 15: Markov blanket of the chance node, $Mod[M_{j,l-1}^t]$, shown in bold.

The joint distribution over the Markov blanket is:

$$
\begin{aligned}
Pr(s^t, a_j^t, o_j^{t+1}, m_{j,l-1}^{t+1}) &= \sum_p Pr(s^t, \mathcal{M}_{j,l-1}^{t,p}, a_j^t, o_j^{t+1}, m_{j,l-1}^{t+1}) \\
&= \sum_p Pr(s^t) Pr(\mathcal{M}_{j,l-1}^{t,p}|s^t) Pr(a_j^t|\mathcal{M}_{j,l-1}^{t,p}) Pr(m_{j,l-1}^{t+1}|\mathcal{M}_{j,l-1}^{t,p}, a_j^t, o_j^{t+1}) Pr(o_j^{t+1}|a_j^t) \\
&= Pr(s^t) Pr(o_j^{t+1}|a_j^t) \sum_p Pr(\mathcal{M}_{j,l-1}^{t,p}|s^t) Pr(a_j^t|\mathcal{M}_{j,l-1}^{t,p}) Pr(m_{j,l-1}^{t+1}|\mathcal{M}_{j,l-1}^{t,p}, a_j^t, o_j^{t+1}) \\
&= Pr(s^t) Pr(o_j^{t+1}|a_j^t) \sum_p \sum_{m_{j,l-1}^t \in \mathcal{M}_{j,l-1}^{t,p}} Pr(m_{j,l-1}^t|s^t) Pr(a_j^t|\mathcal{M}_{j,l-1}^{t,p}) \\
&\quad Pr(m_{j,l-1}^{t+1}|\mathcal{M}_{j,l-1}^{t,p}, a_j^t, o_j^{t+1}) \qquad \text{(from Eq. 2)}
\end{aligned}
\tag{3}
$$

---

3. Because we assume that agents' frames do not change, we may remove the arc from $Mod[M_{j,l-1}^t]$ to $O_j^{t+1}$ thus simplifying the blanket.

The joint distribution prior to aggregation of $Mod[M_{j,l-1}^t]$ is:

$$Pr(s^t, a_j^t, o_j^{t+1}, m_{j,l-1}^{t+1}) = Pr(s^t)Pr(o_j^{t+1}|a_j^t) \sum_{m_{j,l-1}^t \in \mathcal{M}_{j,l-1}^t} Pr(m_{j,l-1}^t|s^t)Pr(a_j^t|m_{j,l-1}^t)$$
$$Pr(m_{j,l-1}^{t+1}|m_{j,l-1}^t, a_j^t, o_j^{t+1})$$

(4)

We equate the right hand sides of Eqs. 3 and 4 to obtain the constraint that must be satisfied by the CPTs of some of the chance nodes in the Markov blanket in order for the joint distribution to remain unchanged:

$$\sum_p Pr(a_j^t|\mathcal{M}_{j,l-1}^{t,p})Pr(m_{j,l-1}^{t+1}|\mathcal{M}_{j,l-1}^{t,p}, a_j^t, o_j^{t+1}) \sum_{m_{j,l-1}^t \in \mathcal{M}_{j,l-1}^{t,p}} Pr(m_{j,l-1}^t|s^t) =$$
$$\sum_{m_{j,l-1}^t \in \mathcal{M}_{j,l-1}^t} Pr(m_{j,l-1}^t|s^t)Pr(a_j^t|m_{j,l-1}^t)Pr(m_{j,l-1}^{t+1}|m_{j,l-1}^t, a_j^t, o_j^{t+1})$$

(5)

Notice that Eq. 5 imposes a constraint on the CPTs of the successor nodes, $A_j^t$ and $Mod[M_{j,l-1}^{t+1}]$. If the constraint is satisfied – a setting for the CPTs of the nodes, $A_j^t$ and $Mod[M_{j,l-1}^{t+1}]$, is found – grouping of AE models in the initial model node is exact and the optimality of the I-DID is preserved. An obvious way to satisfy Eq. 5 would be to meet the following intuitive constraint for each AE class $p$:

$$Pr(a_j^t|\mathcal{M}_{j,l-1}^{t,p})Pr(m_{j,l-1}^{t+1}|\mathcal{M}_{j,l-1}^{t,p}, a_j^t, o_j^{t+1}) =$$
$$\frac{\sum_{m_{j,l-1}^t \in \mathcal{M}_{j,l-1}^{t,p}} Pr(m_{j,l-1}^t|s^t)Pr(a_j^t|m_{j,l-1}^t)Pr(m_{j,l-1}^{t+1}|m_{j,l-1}^t, a_j^t, o_j^{t+1})}{\sum_{m_{j,l-1}^t \in \mathcal{M}_{j,l-1}^{t,p}} Pr(m_{j,l-1}^t|s^t)}$$

(6)

$Pr(a_j^t|m_{j,l-1}^t)$ is fixed for each model, $m_{j,l-1}^t$, in AE class $M_{j,l-1}^{t,p}$ and equals $Pr(a_j^t|\mathcal{M}_{j,l-1}^{t,p})$. Therefore, Eq. 6 reduces to:

$$Pr(m_{j,l-1}^{t+1}|\mathcal{M}_{j,l-1}^{t,p}, a_j^t, o_j^{t+1}) = \frac{\sum_{m_{j,l-1}^t \in \mathcal{M}_{j,l-1}^{t,p}} Pr(m_{j,l-1}^t|s^t)Pr(m_{j,l-1}^{t+1}|m_{j,l-1}^t, a_j^t, o_j^{t+1})}{\sum_{m_{j,l-1}^t \in \mathcal{M}_{j,l-1}^{t,p}} Pr(m_{j,l-1}^t|s^t)}$$

(7)

Observe that Eq. 7 must hold for all values of the physical state, $s^t$. If the right hand side of the above equation remains unchanged for any value of $s^t$, we may set the CPT of $Mod[M_{j,l-1}^{t+1}]$ using it. Typically, it is not trivial – often not possible – to find a single CPT for the chance node $Mod[M_{j,l-1}^{t+1}]$ that will satisfy the constraint for all $s^t$. Chang and Fung (1991) demonstrate that a close approximation would be to take the average of the right hand side of Eq. 7 over all possible values of $s^t$ if these values are close.

Of course, we may wish to aggregate models in node, $Mod[M_{j,l-1}^{t+1}]$, as well (and so on). While the overall procedure is analogous, the difference is in the Markov blanket of the node that is to be aggregated. It includes the predecessor chance nodes, $Mod[M_{j,l-1}^t]$, $A_j^t$, and $O_j^{t+1}$ in addition to its successors and parents of successors corresponding to those in Fig. 15.

We illustrate the application of Eq. 7 to the example policy graph in Fig. 14(a) below:

**Example 2** (Model update). *For simplicity, let the left AE class, $\mathcal{M}_{j,l-1}^{t=0,1}$, comprise of three models, $m_{j,l-1}^{t=0,1}$, $m_{j,l-1}^{t=0,2}$ and $m_{j,l-1}^{t=0,3}$, all of which prescribe action L. Let i's belief over these three models* *be 0.16, 0.23 and 0.32 given physical state TL or TR, respectively (see Fig. 14(a)). We set the probability of updating say, $\mathcal{M}_{j,l-1}^{t=0,1}$, using different action-observation combinations to individual models at time $t$=1, using Eq. 7. We show these probabilities in Fig. 14(b); these form the CPT of node $Mod[M_{j,l-1}^{t=1}]$. Because $Pr(m_{j,l-1}^{t=0,1}|s)$ remains same given any $s$, constraint of Eq. 7 is met and the AE based partitioning at $t$=0 is exact.*

*Next, we group AE models at $t$=1 forming 3 AE classes as shown in Fig. 14(c). Again, we may set the probabilities of updating an AE class given action-observation combinations to individual models at $t$=2 using the right hand side of Eq. 7. However, doing so does not meet the constraint represented by Eq. 7 because for example, $Pr(m_{j,l-1}^{t=1,1}|\mathcal{M}_{j,l-1}^{t=0,p}, a_j^{t=0}, o_j^{t=1})$ varies given different values of the conditionals. As an aside, it is not possible to meet this constraint in this example. Consequently, we adjust the CPT of the chance node, $Mod[M_{j,l-1}^{t=2}]$, according to the average of the right hand side of Eq. 7 for different values of the conditional variables (see Fig. 14(c)). As a result, the AE based partitioning at $t$=1 is not exact.*

A manifestation of the approximation is that agent $i$ may now think that $j$ could initially open the right door, followed by listening and then open the left or right door again. Such a sequence of actions by $j$ was not possible in the original policy graph shown in Fig. 14(a).

## 5.3 Algorithm

We provide an algorithm for exploiting AE in order to solve a level $l \geq 1$ I-DID (as well as a level 0 DID) in Fig. 16. The algorithm starts by selectively solving lower-level I-DID or DID models at $t = 0$, which results in a set of policy trees (line 2). We then build the policy graph by merging the policy trees as mentioned in lines 1-13 of Fig 13. The algorithm differs from Fig. 13 in the expansion phase. In particular, we begin by grouping together AE models in the initial model node. This changes the value of the initial $Mod$ node to the AE classes (lines 3-9). Subsequently, updated models that are AE are aggregated at all time steps, and the CPTs of the $Mod$ nodes are revised to reflect the constraint involving AE classes (lines 13-24). As mentioned in Section 5.2, AE partitioning becomes inexact if we cannot find a CPT for the successor $Mod$ node that satisfies Eq. 7. Given the expanded I-DID, we use the standard look-ahead and backup method to get the solution.

## 5.4 Computational Savings

As we mentioned, the complexity of exactly solving a level $l$ I-DID is, in part, due to solving the lower-level models of the other agent, and given the solutions, due to the exponentially growing space of models. In particular, at some time step $t$, there could be at most $|\mathcal{M}_{j,l-1}^0|(|A_j||\Omega_j|)^t$ many models, where $\mathcal{M}_{j,l-1}^0$ is the set of initial models of the other agent. While $K \ll |\mathcal{M}_{j,l-1}^0|$ models are solved, considering AE bounds the model space to at most $|A_j|$ distinct classes. Thus, the cardinality of the interactive state space in the I-DID is bounded by $|S||A_j|$ elements at any time step. This is a significant reduction in the size of the state space. In doing so, we additionally incur the computational cost of merging the policy trees, which is $\mathcal{O}((|\Omega_j|^{T-1})^2|\mathcal{M}_{j,l-1}^0|^2)$ (from Proposition 2). We point out that this approach is applied recursively to solve I-DIDs at all levels down to 1, as shown in the algorithm.

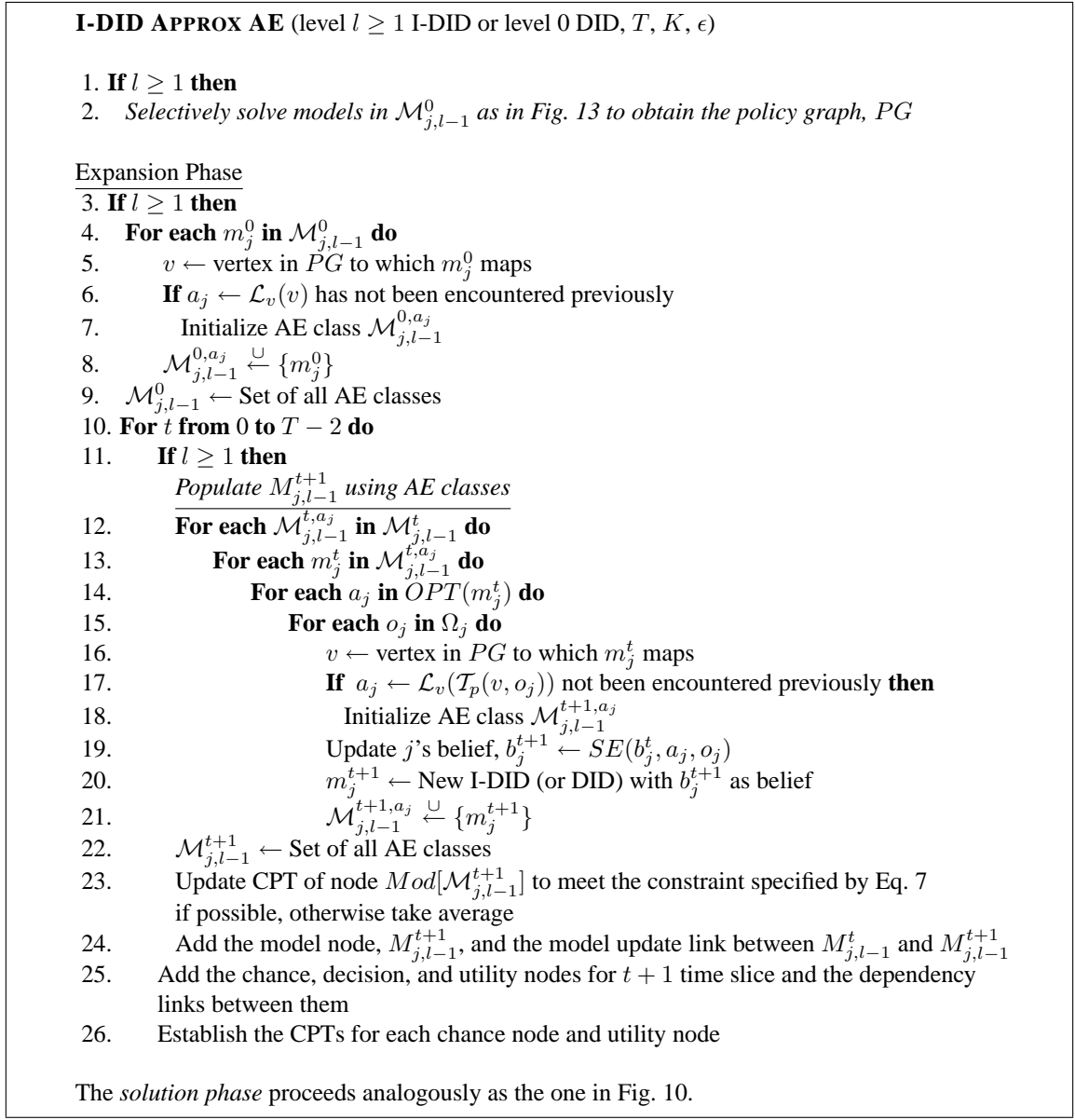

Figure 16: Algorithm for possibly inexactly solving a level $l \geq 1$ I-DID using action equivalence.

## 6. Empirical Results

We implemented the algorithms in Figs. 10, 13 and 16 and refer to the resulting techniques as **Exact-BE**, **DMU** and **AE**, respectively. In addition to these, we utilize the previous approximation technique of $k$-means clustering (Zeng et al., 2007), referred to as **MC**, and the exact approach without exploiting BE, referred to as **Exact**, as baselines. In MC, models are clustered based on the spatial closeness of their beliefs, and the clusters are refined iteratively until they stabilize. Because I-DIDs eventually transform to flat DIDs, we implemented them as a layer above the popular ID tool, HUGIN EXPERT V7.0. The transformed flat DIDs and level 0 DIDs were all solved using HUGIN to obtain the policy trees.

As benchmark problem domains, we evaluate the techniques on two well-known toy problems and a new scalable multiagent testbed with practical implications. One of the benchmarks is the two-agent generalization of the single agent tiger problem introduced previously in Section 3. As we mentioned, our formulation of this problem ($|S|$=2, $|A_i|$=$|A_j|$=3, $|\Omega_i|$=6, $|\Omega_j|$=2) follows the one introduced by Gmytrasiewicz and Doshi (2005), which differs from the formulation of Nair et al. (2003), in not being cooperative and having *door creaks* as additional observations. These observations are informative, though not perfectly, of $j$'s actions. The other toy domain is a generalization of Smallwood and Sondik's machine maintenance problem (Smallwood & Sondik, 1973) to the two-agent domain. This problem ($|S|$=3, $|A_i|$=$|A_j|$=4, $|\Omega_i|$=$|\Omega_j|$=2) is fully described in Appendix B.2. I-DIDs for both these problem domains are shown in Section 3 and Appendix B.2, respectively. Decentralized POMDP solution techniques are not appropriate as baselines in cooperative problems such as machine maintenance because of the absence of a common initial belief among agents, and I-DIDs take the perspective of an agent in the interaction instead of computing the joint behavior.

While the physical dimensions of these problems are small, the interactive state space that includes models of the other agent is an order of magnitude larger. Furthermore, they provide the advantage of facilitating detailed analysis of the solutions and uncovering interesting behaviors as previously demonstrated (Doshi et al., 2009). However, beyond increasing horizons, they do not allow an evaluation of the scalability of the techniques. In this context, we also evaluate the approaches within the Georgia testbed for autonomous control of vehicles (GaTAC) (Doshi & Sonu, 2010), which is a computer simulation framework for evaluating autonomous control of aerial robotic vehicles such as UAVs. Unmanned agents such as UAVs are used in fighting forest fires (Casbeer, Beard, McLain, Sai-Ming, & Mehra, 2005), law enforcement (Murphy & Cycon, 1998), and wartime reconnaissance. They operate in environments characterized by multiple parameters that affect their decisions, including other agents with common or antagonistic preferences. The task is further complicated as the vehicles may possess noisy sensors and unreliable actuators. GaTAC provides a low-cost and open-source alternative to highly complex and expensive simulation infrastructures.

We setup and execute experiments to evaluate the following: ($a$) We hypothesize that in sets of models attributed to the other agent, several are BE. This will lead to the exact approach that groups BE models (Exact-BE) being significantly more efficient than the plain approach (Exact). ($b$) Both approximation techniques (DMU and AE) will improve on the previous approximation technique of $k$-means clustering (MC). This is because MC generates all models at each time step before clustering and furthermore MC may retain BE models. ($c$) Finally, between DMU and AE, we hypothesize AE to be significantly more efficient because it forms as many classes as there are actions only. However, the solution quality resulting from DMU and AE will be explored.

## 6.1 Improved Efficiency Due to BE

We report on the performance of the exact methods (Exact-BE and Exact) when used for solving both level 1 and 2 I-DIDs formulated for the small problem domains. As there are infinitely many computable models, we obtain the policy by *exactly* solving the I-DID given a finite set of models of the other agent initially, $M^0$. In Fig. 17, we show the average rewards gathered by executing the policy trees obtained from exactly solving both level 1 and 2 I-DIDs for the two problem domains, as a function of time allocated toward their solutions.

Each data point is the average of 200 runs of executing the policies, where the true model of the other agent, $j$, is randomly selected according to $i$'s belief distribution over $j$'s models. The time consumed is a function of the initial number of models and the horizon of the I-DID, both of which are varied beginning with $M^0 = 50$ at each level.

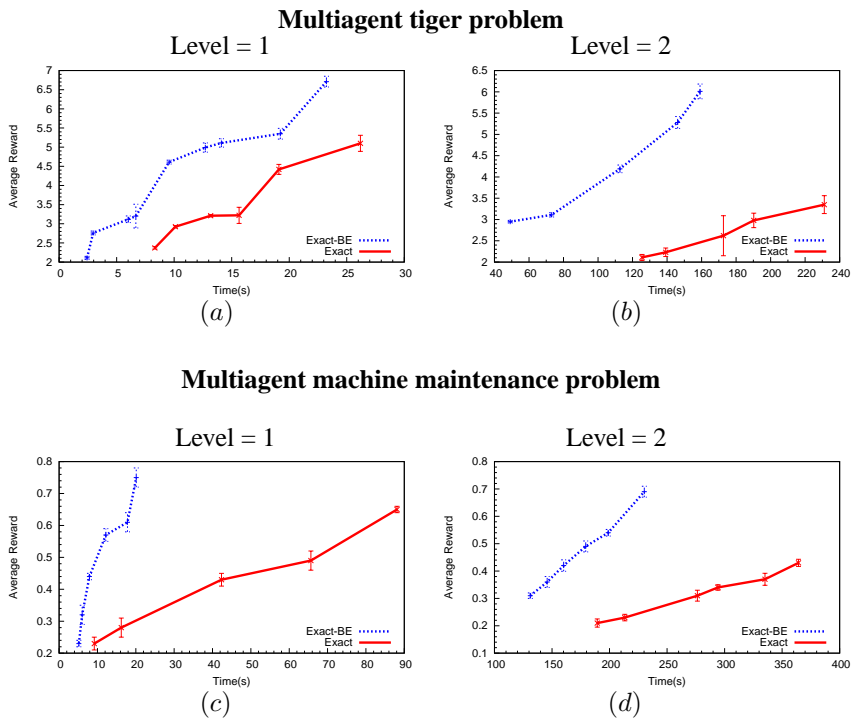

Figure 17: Performance profiles of the exact solutions on the multiagent tiger $(a, b)$ and machine maintenance problems $(c, d)$. Higher average reward for given time is better. Exact-BE significantly improves on the plain Exact approach at levels 1 and 2. For longer times the Exact program runs out of memory. Vertical bars represent the standard deviation from the mean.

From Fig. 17, we observe that Exact-BE performs significantly better than the Exact approach. Specifically, Exact-BE obtains the same amount of reward as Exact but in less time, and subsequently, for a given allocated time, it is able to obtain larger reward than Exact. This is because it is able to solve for a better quality solution in less time as it groups together BE models and retains a single representative from each class, thereby reducing the number of models held in each model node. We see significant improvement in performances in solving I-DIDs at both levels.

We uncover the main reason behind the improved performance of Exact-BE in Fig. 18. As we may expect, after grouping of BE models Exact-BE maintains much fewer classes of models (predicting a particular behavior for the other agent) than the number of individual models maintained by Exact. This occurs for all horizons and for both the problem domains. The number of models increase as horizon reduces (but time steps increase) due to model updates; however, BE classes reduce with smaller horizon because of less distinct behaviors in the solutions.

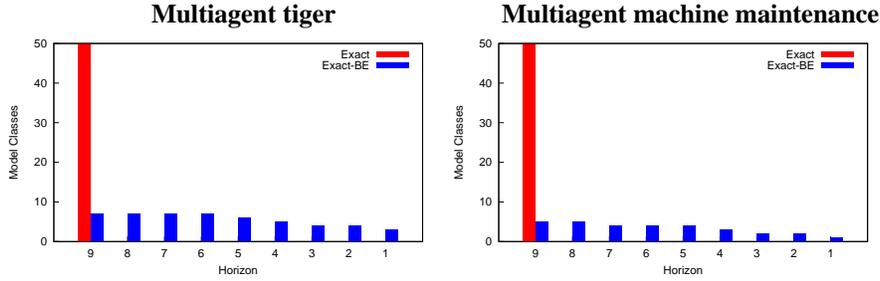

Figure 18: Exact-BE maintains far fewer classes at larger horizons in comparison to the Exact approach. Notice that the number of BE classes reduces as horizon decreases because solutions tend to involve fewer distinct behaviors. We do not show the increase in models with time for Exact for clarity.

As we increase the number of levels beyond two and model $j$ more strategically, we expect Exact-BE (and Exact) to result in solutions whose average reward possibly improves but doesn't deteriorate. However, as the number of models increases exponentially with $l$, we expect substantially more computational resources to be consumed making it challenging to solve for deeper levels.

## 6.2 Comparative Performance of Approximation Methods

While discriminating between model updates as described in Section 4 by itself does not lead to a loss in optimality, we combined it with the approach of solving $K$ (which we will now call $K_{DMU}$) models out of the $M^0$ models, and then solving those models which are not $\epsilon$-close to any of the $K_{DMU}$ models, to form the policy graph. Thus, we initially examine the behavior of the two parameters, $K_{DMU}$ and $\epsilon$, in how they regulate the performance of the DMU-based approximation technique for solving level 1 I-DIDs.

We show the performance of DMU for both the multiagent tiger and machine maintenance problems in Fig. 19. We also compare its performance with an implementation of MC; $K_{MC}$ represents the total number of models retained after clustering and pruning in the approach. The performance of MC is shown as flat lines because $\epsilon$ does not play any role in the approach. Each data point for DMU is the average of 50 runs of executing the policies where the true model of the other agent, $j$, is randomly picked according to $i$'s belief distribution over $j$'s models, and solved exactly if possible. Otherwise, if $l > 1$, we solve it approximately using DMU with a large $K_{DMU}$ and small $\epsilon$. The plot is for $M^0 = 100$, and a horizon of 10. As we increase the number of models randomly selected, $K_{DMU}$, and reduce distance, $\epsilon$, the policies improve and converge toward the exact. Notice that DMU improves on the performance of MC as we reduce $\epsilon$, for $K_{DMU} = K_{MC}$. This behavior remains true for the multiagent machine maintenance problem as well.

We evaluate the impact of AE on solving level 1 I-DIDs for both problem domains and compare it to MC. The experiments were run analogously as before with different values for $K_{AE}$ and $\epsilon$. Both these parameters play roles that are similar to their use in DMU. We observe from Fig. 20 that as $\epsilon$ reduces and more models beyond $K_{AE}$ are solved, the solution generated by AE improves on MC for the case where $K_{AE} = K_{MC}$. Of course, as we solve more models initially, AE produces better quality solutions because the generated policy graph includes more parts of the exact graph. Additionally, as we may expect, the solution quality approaches that of the exact and becomes significantly close to the exact (within one standard deviation for the tiger problem) for $\epsilon < 0.1$.

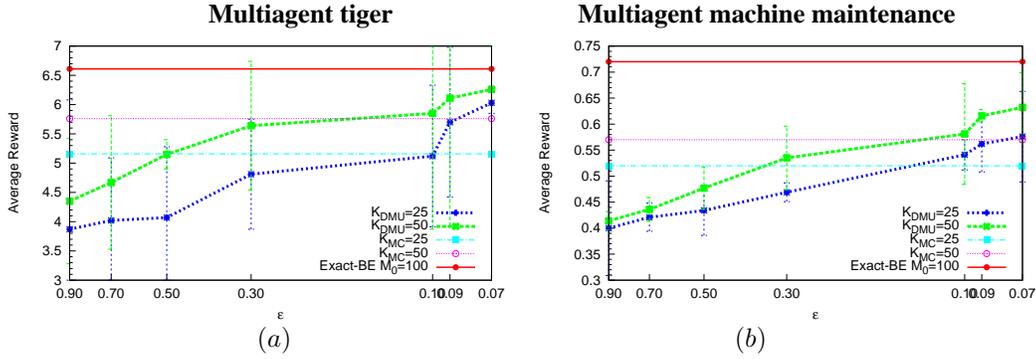

Figure 19: Performance profiles for DMU on horizon $T$=10, level 1 I-DID with $M^0$=100. For given $M^0$, the performance approaches that of the exact method as $K_{DMU}$ increases and $\epsilon$ reduces. The comparison with MC indicates that better performance is achieved by DMU-based solutions.

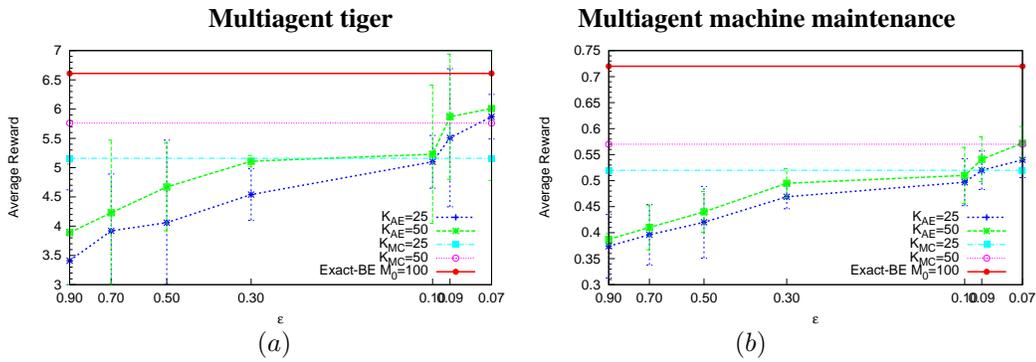

Figure 20: Performance profiles for AE on horizon $T$=10, level 1 I-DID with $M^0$=100. For given $M^0$, the performance approaches that of the exact method as $K_{AE}$ increases and $\epsilon$ reduces. Comparative performance in relation to MC indicates that AE is capable of achieving better quality solutions (although for relatively small values of $\epsilon$).

While our previous experiments demonstrated that both DMU and AE are capable of improving on MC, it is not clear how many more initial models beyond $K_{MC}$ were solved to obtain the improvements. Furthermore, performance of DMU and AE were not compared. In Fig. 21, we directly compare the performance of DMU, AE and MC. In particular, we measure the average rewards obtained by corresponding solutions of level 2 I-DIDs as a function of time consumed by the approaches. For DMU and AE, the time taken is dependent on the parameters ($K$ and $\epsilon$), tree merging and the horizon of the I-DID solved. For MC, the time is due to the iterative clustering until convergence and the $K$ models that are picked. For the problem domains considered, larger horizon with increasing $K$ and reducing $\epsilon$ typically leads to better average rewards. We observe that both DMU and AE significantly improve on MC – they produce identical quality solutions in less time than that by MC. Furthermore, between DMU and AE, the latter's performance is more favorable. This is because, while grouping AE models may result in an additional approximation, further efficiency is made possible by the fewer AE classes and whose number does not exceed a constant across horizons.

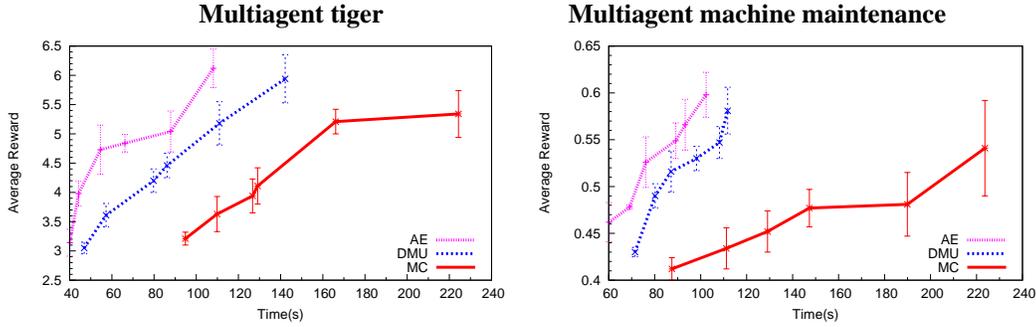

Figure 21: Performance comparison of approximation techniques for solving level 2 I-DIDs. AE achieves significantly improved efficiency for identical quality solutions over the two domains.

We empirically explore the reason behind the comparative performance of the approximation techniques. Because the time (and space) consumed by the approaches is predominantly due to the solution of the models in the model node, we focus on the models retained by the approaches at different horizons. Fig. 22 shows the models at different horizons for varying $\epsilon$ and for both the problem domains. Note that when $\epsilon = 0$, all initial models are solved and in the case of DMU, results in the behaviorally minimal set at every horizon. Furthermore, as we mentioned previously, for non-zero $\epsilon$, the merged policy graph is a subgraph of the exact ($\epsilon = 0$) case. As we show, the resulting sets of models are subsets of the minimal set.

At any horizon, AE maintains no more model classes than the number of $j$'s actions, $|A_j|$. As we see, this is substantially less than the number maintained by DMU. MC maintains a fixed number, $K_{MC}$, of models at each horizon of the I-DID.

## 6.3 Runtime Comparison

We show the run times of the exact and approximation techniques for solving level 1 and 2 I-DIDs while scaling in horizons, in Table 1. Notice that a plain exact approach that does not exploit model equivalences scales poorly beyond small horizons. In contrast, simply grouping BE models and reducing the exponential growth in models leads to significantly faster executions and better scaleup. The run times are reported for both these approaches solving the same I-DID exactly.

In obtaining the run times for the approximations, we adjusted the corresponding parameters so that the quality of the solution by each approach was similar to each other. DMU and AE reported substantially less execution times and better scaleup in comparison to MC for both the domains. However, the run times of DMU and AE are relatively similar for level 1 I-DIDs. Although, as we saw previously, there is a difference in the number of models maintained by the two techniques, solving $j$'s level 0 DIDs is quick and the difference does not lead to a significant impact. The differences in run times are more significant for level 2 I-DIDs because solving $j$'s level 1 I-DIDs is computationally intensive. As we may expect, AE consumes substantially less time in comparison to DMU – sometimes less than half. Both approaches scale up similarly in terms of the horizon. In particular, we are able to solve both level 1 and 2 I-DIDs for more than a horizon of 10. Further scaleup is limited predominantly due to our use of software such as HUGIN for solving the flat DIDs, which seeks to keep the entire transformed DID in main memory.

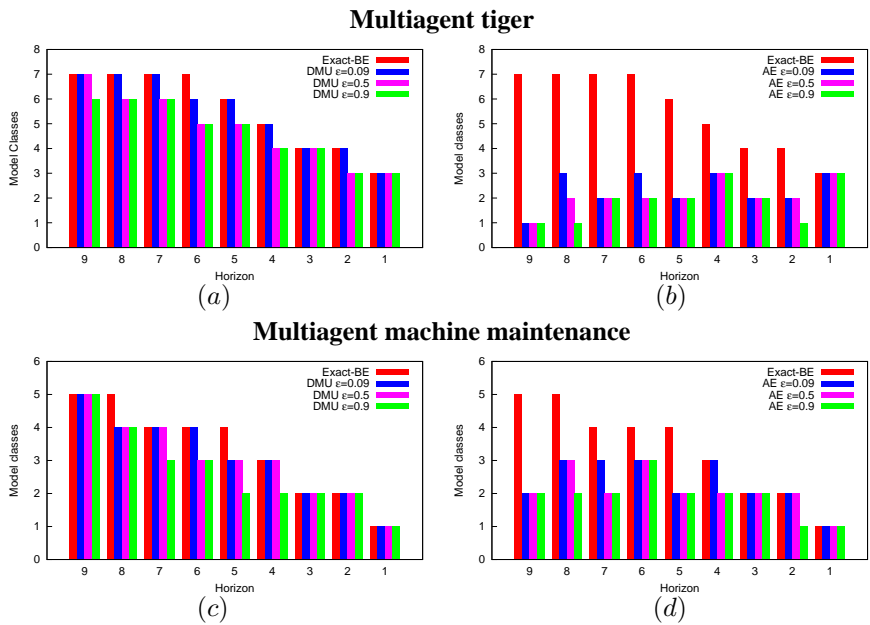

Figure 22: Number of models maintained in the model node for different horizon in a level 1 I-DID for the multiagent tiger ($a$, $b$) and machine maintenance ($c$, $d$) problems. For DMU, as $\epsilon$ reduces, the model space approaches the minimal set. Aggregation using AE further reduces the model space.

### 6.4 Scalable Testbed: GaTAC

As we mentioned, the objective behind developing GaTAC is to provide a realistic and scalable testbed for algorithms on multiagent decision making. GaTAC facilitates this by providing an intuitive and easy to deploy architecture that makes use of powerful, open-source software components. Successful demonstrations of algorithms in GaTAC would not only represent tangible gains but have the potential for practical applications toward designing autonomous vehicles such as UAVs. [4]

---

4. GaTAC is available for download at `http://thinc.cs.uga.edu/thinclabwiki/index.php/GaTAC_:_Georgia_Testbed_for_Autonomous_Control_of_Vehicles`

| Level | Domain | T | Exact | Exact-BE | DMU | AE | MC |
|-------|--------|---|-------|----------|-----|-----|-----|
| Level 1 | Tiger | 4 | 5.8s | 0.47s | 0.13s | 0.42s | 2.12s |
| | | 8 | * | 10.5s | 1.27s | 1.64s | 28.45s |
| | | 14 | * | 2h 4m | 15m 6s | 15m 15s | * |
| | MM | 4 | 6.66s | 0.45s | 0.19s | 0.22s | 3.23s |
| | | 6 | * | 1.73s | 0.53s | 0.58s | 9.88s |
| | | 12 | * | 9m 40s | 2m 9s | 2m 12s | * |
| Level 2 | Tiger | 3 | 3m 24s | 10.97s | 4.63s | 3.11s | 1m 46s |
| | | 6 | * | 22m 6s | 6m 54s | 3m 3s | * |
| | | 10 | * | 2h 48m | 27m 36s | 16m 54s | * |
| | MM | 4 | 5m12s | 1.11s | 0.33s | 0.58s | 2m 21s |
| | | 6 | * | 13.59s | 4.3s | 1.48s | * |
| | | 10 | * | 20m 36s | 3m 36s | 2m 15s | * |

Table 1: Exploiting model equivalences has a significant impact on the execution times. Both DMU and AE demonstrate improved efficiency. Algorithm involving AE scales significantly better to larger horizons for deeper strategy levels. All experiments are run on a WinXP platform with a dual processor Xeon 2.0GHz and 2GB memory. '*' indicates that the data point is unavailable because the program ran out of memory.

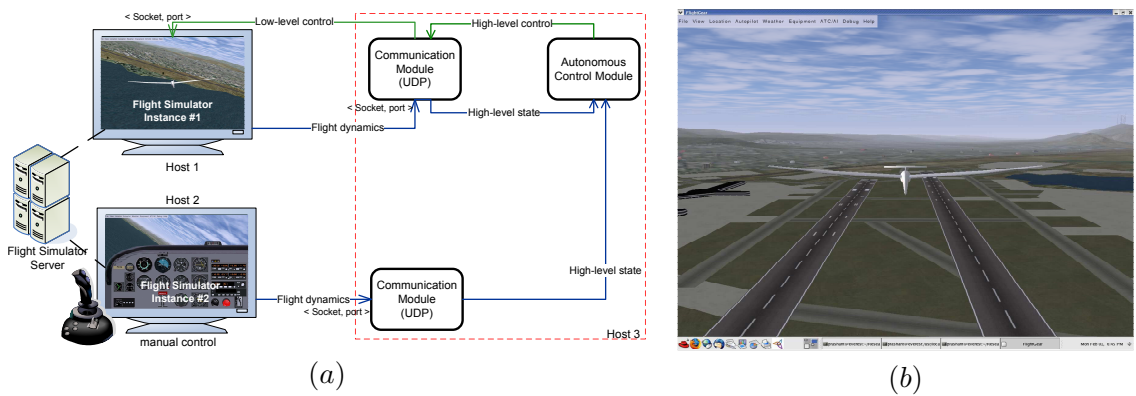

(a)                                                                    (b)

Figure 23: (a) Design of GaTAC showing two networked instances of a flight simulator (FlightGear with 3D scenery from TerraGear), one autonomously and other manually controlled. GaTAC is extensible and more instances may be added. (b) Snapshot of a UAV flying within FlightGear. Different viewpoints including an external view as shown and a cockpit view are available.

A simplified design of the GaTAC architecture is shown in Fig. 23, where a manually controlled UAV is interacting with an autonomous one. Briefly, GaTAC employs multiple instances of an open-source flight simulator, called *FlightGear* (Perry, 2004), possibly on different networked platforms that communicate with each other via external servers, and an autonomous control module that interacts with the simulator instances. GaTAC can be deployed on most platforms including Linux and Windows with moderate hardware requirements, and the entire source code is available under GNU Affero public license version 3.

We utilize a relatively straightforward setting consisting of another hostile fugitive, who is the target of ground reconnaissance (Fig. 24). UAV $I$ must track down the fugitive before he flees to the safe house. The problem is made complex by assuming that the fugitive is unaware of its own precise location though it knows the location of the safe house, and $I$ may not be aware of the fugitive's location. The problem is further complicated if we realistically assume nondeterministic actions and observations. Our simulations in GaTAC include grids of sizes $3 \times 3$ and $5 \times 5$ with the same actors in each. GaTAC may be programmed to support more complex scenarios comprising of a team of UAVs, multiple hostile UAVs and reconnaissance targets attempting to blend in with civilians.

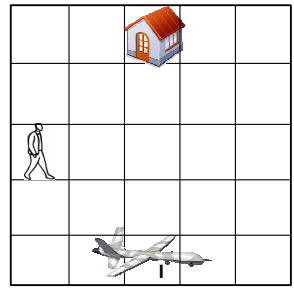

Figure 24: Example $5 \times 5$ theater of UAV $I$, which performs low-altitude reconnaissance of a potentially hostile theater populated by a fugitive, $J$.

We summarize our formulation of the UAV's problem domain. We utilize the possible relative positions of the fugitive as states. Hence, possible states would be *same*, *north*, *south*, *east*, *west*, *north-west*, and so on. Our representation for a $3 \times 3$ theater consists of 25 physical states for the UAV $I$. We assume that the fugitive is unaware of its own location resulting in 9 physical states for it. Extending the theater to a $5 \times 5$ grid leads to 81 physical states for the UAV and 25 for the fugitive. We factor the physical state into two variables in the I-DID that model the row and column positions, respectively. Both UAV $I$ and the fugitive may move in one of the four cardinal directions, and they may additionally hover at their current positions and listen to get informative observations. Thus the *actions* for both $I$ and the fugitive are {*move_north, move_south, move_west, move_east, listen*}. We may synchronize the actions for the two agents in GaTAC by allocating equal time duration to the performance of each action. Typically, UAVs have infrared and camera sensors whose range is limited. Accordingly, we assume that both the UAV $I$ and the fugitive can sense whether their respective target is north of them (*sense_north*), south of them (*sense_south*), west or east of them in the same row (*sense_level*) or in the same location as them (*sense_found*). For $I$ the target is the fugitive, while the fugitive's target is the safe house. If we assume that the fugitive is unaware of $I$'s presence, its *transition function* is straightforward and simply reflects the possible nondeterministic change in grid location of the fugitive as it moves or listens. However, transitions in physical state of $I$ are contingent on the joint actions of both agents. Furthermore, the probability distribution over the next states is not only due to the nondeterminism of the actions, but is also influenced by the current relative physical state. To provide an opportunity for the UAV $I$ to catch the fugitive, we assume that the fugitive can sense the safe house only when it is within a distance of 1 sector (horizontally or vertically) from it. On the other hand, UAV $I$'s observations of the fugitive are not limited by this constraint. Thus, if the fugitive is in any location that is north of $I$ (including north-west or north-east), $I$ receives an observation of *sense_north*. To simulate noise in the sensors, we assume that the likelihood of the correct observation is 0.8 while all others are equiprobable. The reward function is straightforward with the fugitive receiving a reward if its location is identical to that of the safe house, and small costs for performing actions to discourage excessive action taking. Analogously, UAV $I$ recieves a reward on performing an action and receiving an observation of *sense_found*, and incurs small costs for actions that lead to other observations.

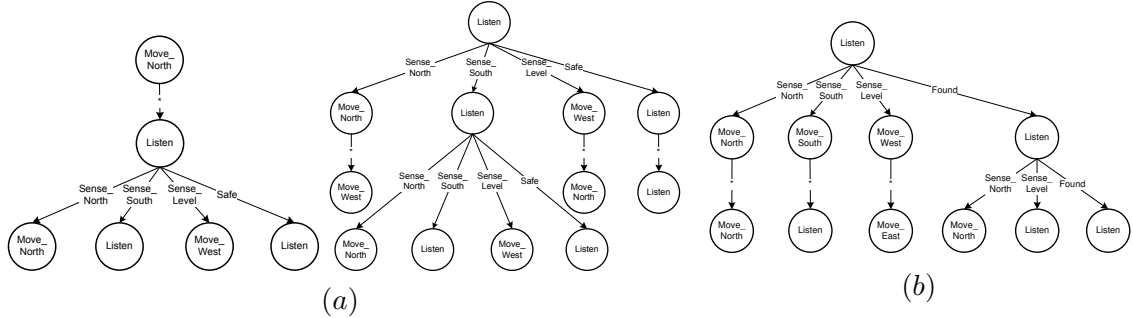

Figure 25: $(a)$ Example policies of the fugitive modeled by the UAV. $(b)$ UAV $I$'s optimal policy of pursuing the fugitive obtained by solving the level 1 I-DID exactly using BE. The policy is straightforward, using observations to guide the actions. All policies are for a $3 \times 3$ grid.

We modeled the problem formulation described above using a level 1 I-DID for the UAV $I$ and level 0 models for the fugitive. We show two example policies of the fugitive obtained by solving its level 0 models, in Fig. 25$(a)$. While we considered several models for the fugitive with differing initial beliefs, the fugitive's initial belief of likely being just below the safe house results in the left policy, while its initial belief of likely being south and east of the safe house leads to the policy on the right. We show the UAV's policy of reconnaissance in Fig. 25$(b)$, obtained by solving its level 1 I-DID exactly while utilizing BE classes. Thirty models of the fugitive grouped into 16 BE classes were considered in the I-DID. Here, the UAV initially believes that the fugitive is likely to be in the same row or south of it.

We simulate the reconnaissance theaters of Fig. 24 in GaTAC. The UAV and the fugitive's behaviors are controlled by their respective policies provided as input to the autonomous control module. For each simulation run, we generated the UAV's policy by solving its level 1 I-DID using either Exact-BE, DMU or AE, and sampled one of the fugitive's 30 models based on the UAV's initial belief. The run terminates when either the fugitive reaches the safe house, or the UAV spots the fugitive by entering the same sector as the fugitive. In the case of DMU and AE, we used parameters, $K = 13$ and $\epsilon = 0.3$ for the $3 \times 3$ problem size, and $K = 17, \epsilon = 0.15$ for the $5 \times 5$ problem. We show the average reward gathered by the UAV across 20 simulation runs for each of the three approaches in Fig. 26$(a)$ and the associated clock time for solving the I-DIDs in Fig. 26$(b)$. While we considered several different beliefs for the UAV, positioning itself approximately between the fugitive and the safe house yielded a fugitive capture rate of 65% among the simulation runs and an escape rate of 25%. The remaining runs did not result in a capture or an escape.

While exactly solving the I-DID using Exact-BE continues to provide the largest reward among all approaches, as shown in Fig. 26$(a)$, it fails to scale to a longer horizon or problem size. Both DMU and AE scale, although AE performs worse than DMU in the context of reward in this domain. Note that longer horizons result in overall better quality policies in this problem domain, as we may expect. This is because the UAV is able to plan its initial action better. Finally, the improved reward obtained by AE relative to DMU's as the horizon increases from 6 to 8 for the $3 \times 3$ grid, and

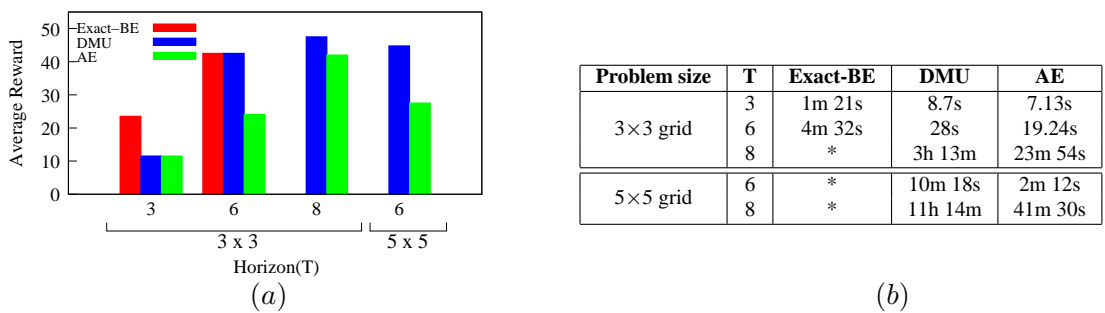

$(a)$                                                           $(b)$

Figure 26: $(a)$ Simulation performance of the UAV in GaTAC when its level 1 I-DID is solved using the different approaches and over longer horizons. We scaled the grid size from $3 \times 3$ to $5 \times 5$. Notice that Exact-BE fails to scale as the horizon and problem size increase and we do not show its reward. $(b)$ Execution times for solving the level 1 I-DID using the different approaches. Although AE results in solutions of lower quality compared to DMU, it does so in much less time. For the longer horizon of 8, this time difference is about an order of magnitude.

the slight climb in AE's relative reward as a percentage of DMU's from about 57% to 62% for the horizon of 6 as the grid size is scaled from $3 \times 3$ to $5 \times 5$ makes us believe that AE's performance will not necessarily deteriorate compared to DMU for larger problems. Overall, we demonstrate the scalability of DMU and AE by increasing the horizon to 8 for this larger problem domain, and further scaling its size. Larger grid sizes or longer horizons resulted in DIDs that could not be solved by HUGIN given the fixed memory.

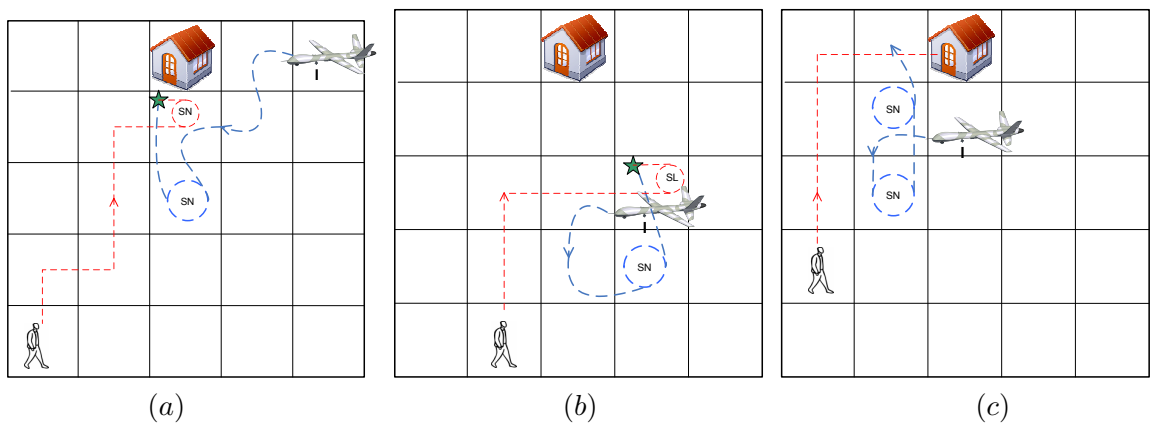

$(a)$                                   $(b)$                                   $(c)$

Figure 27: UAV $I$'s flight trajectory is in dashed blue while the fugitive's is in dashed red. Trajectories $(a, b)$ eventually lead to the fugitive being spotted while in $(c)$, the fugitive reaches the safe house. The latter is due to an incorrect move by the UAV because of ambiguity in its observations. A circle represents hovering during which the UAV or the fugitive is listening and senses the labeled observation.

Finally, we handpick three simulations from the numerous that we carried out and show the corresponding trajectories of the UAV and the fugitive in Fig. 27. We show two trajectories where the UAV spots the fugitive and a trajectory where the fugitive successfully escapes to the safehouse. Figure 27 shows that the trajectories of the UAV can get quite complicated while those of the fugitive are more straightforward due to its low strategic awareness.

## 7. Discussion

Graphical models are an appealing formalism for modeling decision making due to the convenience of representation and the increased efficiency of the solution. DIDs are a key contribution in this regard. I-DIDs are founded on the normative paradigm of decision theory as formalized by DIDs and augmented with aspects of Bayesian games (Harsanyi, 1967) and interactive epistemology (Aumann, 1999a, 1999b) to make them applicable to interactions. I-DIDs generalize DIDs to multiagent settings thereby extending the advantages of DIDs to decision making in multiagent settings. I-DIDs adopt a subjective approach to understanding strategic behavior, rooted in a decision-theoretic formalism that takes a decision-makers perspective in an interaction which may be cooperative or non-cooperative. The broader impact is that understanding the agent's decision-making process facilitates planning and problem-solving at its own level and in the absence of centralized controllers or assumptions about agent behaviors. In a game-theoretic sense, the setting modeled by I-DIDs is a partially observable stochastic game, and solving it by computing Nash equilibria or otherwise, has received minimal attention in game theory.

We presented a collection of exact and approximation algorithms for scalably solving I-DIDs. These algorithms improve on early techniques by providing more effective approaches in order to reduce the exponential growth of other agents' models at each time step. Our main idea is to cluster models attributed to other agents that are BE. These models attribute identical behaviors across all time steps to the other agent. We then select representative models from each cluster without loss of optimality in the solution. Instead of generating the updated models and clustering them, we showed how we may selectively update those models that will not be BE to existing models in the next time step. Nevertheless, ascertaining BE requires solving the initial set of models. In order to approximate this, we proposed solving $K$ randomly picked models followed by all those which are not $\epsilon$-close to any of the $K$ models. We partially bounded the error due to this approximation for some cases. Despite the lack of a proper bound, our empirical results reveal that the error becomes unwieldy for large $\epsilon$ values only. This is because many problems admit large BE regions for models, albeit which tend to reduce as horizon increases.

In order to further reduce the number of equivalence classes, we investigated grouping together models whose prescribed actions at a particular time step are identical. This approach is appealing because the number of AE classes is upper bounded by the number of distinct actions at any time. While AE models may be grouped without loss in optimality, we identified the conditions under which AE leads to an approximation. Our experiments indicate that considerations of BE are of significance while grouping AE models leads to most reduction in the model space among all the different approaches. However, they also show that the gap in the quality of the solutions due to grouping BE and grouping AE models can become large. This difference depends on the domain characteristics and does not necessarily worsen as the problem is scaled in horizon or size.

Due to the expressiveness of the modeling and its ensuing complexity, our experimentation focused on settings involving two agents. However, as the number of other agents increases, po-tential computational savings due to exploiting BE and AE assumes greater significance. This is because the space of interactive states increases exponentially with the number of agents. Therefore, grouping agent models in less numbers of classes would substantially reduce the state space and consequently the size of the I-DID. We may group models separately for each agent in which case the computations for ascertaining BE classes grow linearly with the number of agents. On the other hand, consider the tiger problem where agent $i$'s action is affected by somebody opening the door, without the need for knowing which particular agent opened it. In this case, we may group together models that are BE but belonging to different agents, leading to increased savings. Our preliminary experimentation in the context of the multiagent tiger problem in a setting involving two other agents (total of three agents) one of which is thought to be cooperative while the other adversarial, indicates that grouping BE models for each other agent leads to a speed up of about 7 for both a 3-horizon and 5-horizon I-DID. Specifically, the computation time reduces from 4.8s for Exact to 0.6s for Exact-BE when the horizon is 3, and from 72.6s to 10.5s for Exact-BE when the horizon is 5.

Because identifying exact BE is computationally intensive, we think that improved scalability may be brought by further investigating approximate BE of models. This would allow us to form larger clusters and have fewer representative models. While we are investigating multiple ways of doing this, a significant challenge is storing policy trees that grow exponentially as the horizon is further scaled. One promising approach in this regard is to compare partial policy trees of bounded depth and the distance between the belief vectors at the leaves of the trees. This allows us to define an approximate measure of BE based on the distance between the updated belief vectors given that the bounded-depth policy trees are identical. However, our preliminary investigations reveal that deriving the depth of the tree becomes challenging for certain types of problems.

A general limitation of utilizing the spatial closeness of beliefs for approximately identifying BE models is that the error may be larger if the frames in the models differ. This is because model beliefs that are close are still less likely to result in the same behavior if say, the reward functions are different. In the absence of this approximation, our approaches for discriminatively updating models and grouping AE models continue to apply if frames are also uncertain because they operate on model solutions – policy trees and actions – and not on model specifications. Another impact of considering frame uncertainty is that the Markov blanket shown in Fig. 15 changes. A general hurdle is that further scalability of ID-based graphical models is also limited by the absence of state-of-the-art techniques for solving DIDs within commercial implementations such as HUGIN EXPERT that predominantly rely on solving the entire DID in main memory. Although newer versions of HUGIN allow the use of limited memory IDs (Nilsson & Lauritzen, 2000), recent advances such as a branch-and-bound approach for solving multistage IDs (Yuan, Wu, & Hansen, 2010) would help drive further scalability of I-DID solutions.

## Acknowledgments

Yifeng Zeng acknowledges support from the Obel Family Foundation (Denmark) and NSFC (# 60974089 and # 60975052). Prashant Doshi acknowledges support from an NSF CAREER grant (# IIS-0845036) and a grant from U.S. Air Force (# FA9550-08-1-0429). The authors also thank Ekhlas Sonu and Yingke Chen for help in performing the GaTAC-based simulations, and acknowledge all the anonymous reviewers for their helpful comments.

## Appendix A. Proofs

*Proof of Proposition 1.* We prove by induction on the horizon. Let $\{\mathbb{M}_{j,l-1}^1, \ldots, \mathbb{M}_{j,l-1}^q\}$ be the collection of behaviorally equivalent sets of models in $\mathcal{M}_{j,l-1}$. We aim to show that the value of each of $i$'s actions in the decision nodes at each time step remains unchanged on application of the transformation, $X$. This implies that the solution of the I-DID is preserved. Let $Q^n(b_{i,l}, a_i)$ give the action value at horizon $n$. Its computation in the I-DID could be modeled using the standard dynamic programming approach. Let $ER_i(s, m_{j,l-1}, a_i)$ be the expected immediate reward for agent $i$ averaged over $j$'s predicted actions. Then, $\forall_{m_{j,l-1}^q \in \mathbb{M}_{j,l-1}^q}$ $ER_i(s, m_{j,l-1}^q, a_i) = \sum_{a_j} R_i(s, a_i, a_j)$ $Pr(a_j|m_{j,l-1}^q) = R_i(s, a_i, a_j^q)$, because $a_j^q$ is optimal for all $m_{j,l-1}^q \in \mathbb{M}_{j,l-1}^q$.

**Basis step:** $Q^1(b_{i,l}, a_i) = \sum_{s, m_{j,l-1}} b_{i,l}(s, m_{j,l-1}) ER_i(s, m_{j,l-1}, a_i)$

$= \sum_{s,q} b_{i,l}(s) \sum_{m_{j,l-1}^q \in \mathbb{M}_{j,l-1}^q} b_{i,l}(m_{j,l-1}^q|s) R_i(s, a_i, a_j^q)$   ($a_j^q$ is optimal for all behaviorally equivalent models in $\mathbb{M}_{j,l-1}^q$)

$= \sum_{s,q} b_{i,l}(s) R_i(s, a_i, a_j^q) \sum_{m_{j,l-1}^q \in \mathbb{M}_{j,l-1}^q} b_{i,l}(m_{j,l-1}^q|s)$

$= \sum_{s,q} b_{i,l}(s) R_i(s, a_i, a_j^q) \hat{b}_{i,l}(\hat{m}_{j,l-1}^q|s)$     (from Eq. 1)

$= \sum_{s,q} \hat{b}_{i,l}(s, \hat{m}_{j,l-1}^q) ER_i(s, \hat{m}_{j,l-1}^q, a_i)$   ($a_j^q$ is optimal for representative $\hat{m}_{j,l-1}^q$)

$= \hat{Q}^1(\hat{b}_{i,l}, a_i)$

**Inductive hypothesis:** Let, $\forall_{a_i, b_{i,l}} Q^n(b_{i,l}, a_i) = \hat{Q}^n(\hat{b}_{i,l}, a_i)$, where $\hat{b}_{i,l}$ relates to $b_{i,l}$ using Eq. 1. Therefore, $U^n(b_{i,l}) = \hat{U}^n(\hat{b}_{i,l})$ where $U^n(b_{i,l})$ is the expected utility of $b_{i,l}$ for horizon $n$.

**Inductive proof:** $Q^{n+1}(b_{i,l}, a_i) = \hat{Q}^1(\hat{b}_{i,l}, a_i) + \sum_{o_i, s, m_{j,l-1}, a_j} Pr(o_i|s, a_i, a_j)$ $Pr(a_j|m_{j,l-1}) b_{i,l}(s, m_{j,l-1}) U^n(b_{i,l}')$   (basis step)

$= \hat{Q}^1(\hat{b}_{i,l}, a_i) + \sum_{o_i, s, q} Pr(o_i|s, a_i, a_j^q) b_{i,l}(s) \sum_{m_{j,l-1}^q \in \mathbb{M}_{j,l-1}^q} b_{i,l}(m_{j,l-1}^q|s) U^n(b_{i,l}')$   ($a_j^q$ is optimal for models in $\mathbb{M}_{j,l-1}^q$)

$= \hat{Q}^1(\hat{b}_{i,l}, a_i) + \sum_{o_i, s, q} Pr(o_i|s, a_i, a_j^q) b_{i,l}(s) \sum_{m_{j,l-1}^q \in \mathbb{M}_{j,l-1}^q} b_{i,l}(m_{j,l-1}^q|s) \hat{U}^n(\hat{b}_{i,l}')$   (using the inductive hypothesis)

$= \hat{Q}^1(\hat{b}_{i,l}, a_i) + \sum_{o_i, s, q} Pr(o_i|s, a_i, a_j^q) b_{i,l}(s) \hat{b}_{i,l}(\hat{m}_{j,l-1}^q|s) \hat{U}^n(\hat{b}_{i,l}')$   (from Eq. 1)

$= \hat{Q}^1(\hat{b}_{i,l}, a_i) + \sum_{o_i, s, q} Pr(o_i|s, a_i, a_j^q) \hat{b}_{i,l}(s, \hat{m}_{j,l-1}^q) \hat{U}^n(\hat{b}_{i,l}')$

$= \hat{Q}^{n+1}(\hat{b}_{i,l}, a_i)$                                                                 $\square$

**Calculating prediction error in Section 4.5.** Let $m_{j,l-1}$ be the model associated with a solved model, $m_{j,l-1}'$, resulting in the worst error. Let $\alpha$ be the exact policy tree obtained by solving $m_{j,l-1}$ optimally and $\alpha'$ be the policy tree for $m_{j,l-1}'$. As $m_{j,l-1}'$ is itself solved inexactly due to approximate solutions of lower level models, let $\alpha''$ be the exact policy tree that is optimal for $m_{j,l-1}'$. If $b_{j,l-1}$ is the belief in $m_{j,l-1}$ and $b_{j,l-1}'$ in $m_{j,l-1}'$, then the error is:

$$
\begin{aligned}
E \quad &= |\alpha \cdot b_{j,l-1} - \alpha' \cdot b_{j,l-1}| \\
&= |\alpha \cdot b_{j,l-1} - \alpha' \cdot b_{j,l-1} + (\alpha'' \cdot b_{j,l-1} - \alpha'' \cdot b_{j,l-1})| \quad \text{(add zero)} \\
&= |(\alpha \cdot b_{j,l-1} - \alpha'' \cdot b_{j,l-1}) + (\alpha'' \cdot b_{j,l-1} - \alpha' \cdot b_{j,l-1})| \\
&\leq |(\alpha \cdot b_{j,l-1} - \alpha'' \cdot b_{j,l-1})| + |(\alpha'' \cdot b_{j,l-1} - \alpha' \cdot b_{j,l-1})| \quad \text{(triangle inequality)}
\end{aligned}
\tag{8}
$$

For the first term, $|\alpha \cdot b_{j,l-1} - \alpha'' \cdot b_{j,l-1}|$, which we denote by $\rho$, the error is due to associating $m_{j,l-1}$ with $m_{j,l-1}'$, both solved exactly. We analyze this error below:

$$
\begin{aligned}
\rho \quad &= |\alpha \cdot b_{j,l-1} - \alpha'' \cdot b_{j,l-1}| \\
&= |\alpha \cdot b_{j,l-1} - \alpha'' \cdot b'_{j,l-1} + \alpha'' \cdot b'_{j,l-1} - \alpha'' \cdot b_{j,l-1}| \quad \text{(add zero)} \\
&\leq |\alpha \cdot b_{j,l-1} - \alpha \cdot b'_{j,l-1} + \alpha'' \cdot b'_{j,l-1} - \alpha'' \cdot b_{j,l-1}| \quad (\alpha'' \cdot b'_{j,l-1} \geq \alpha \cdot b'_{j,l-1}) \\
&= |\alpha \cdot (b_{j,l-1} - b'_{j,l-1}) - \alpha'' \cdot (b_{j,l-1} - b'_{j,l-1})| \\
&= |(\alpha - \alpha'') \cdot (b_{j,l-1} - b'_{j,l-1})| \\
&\leq ||\alpha - \alpha''||_\infty \times ||b_{j,l-1} - b'_{j,l-1}||_1 \quad \text{(Hölder's inequality)} \\
&\leq (R_j^{max} - R_j^{min}) T \times \epsilon
\end{aligned}
\tag{9}
$$

In the above inequality, the largest difference between $b_{j,l-1}$ and $b'_{j,l-1}$ is $\epsilon$, otherwise model, $m_{j,l-1}$ with belief $b_{j,l-1}$, would be solved. Notice that the error is regulated by $\epsilon$, and as $\epsilon$ increases, we solve less models beyond $K$ and the approximation error worsens.

In subsequent time steps, because the sets of models could be subsets of the minimal sets, the updated probabilities could be transferred to incorrect models. In the worst case, the error incurred is bounded analogously to Eq. 9. Hence, the cumulative error in $j$'s predicted behavior over $T$ steps is at most $T \times \rho$, which is similar to that of the previous $k$-means model clustering approach (Zeng et al., 2007):

$$
\rho^T \leq (R_j^{max} - R_j^{min}) T^2 \epsilon
$$

The second term, $|(\alpha'' \cdot b_{j,l-1} - \alpha' \cdot b_{j,l-1})|$, in Eq. 8 represents the error due to the approximate solutions of models further down in level (for example, $i$'s level $l-2$ models). Since $j$'s behavior depends, in part, on the actions of $i$ (and not on the value of $i$'s solution), even a slight deviation by $j$ from the exact prediction for $i$ could lead to $j$'s behavior with the worst error. Hence, it seems difficult to derive bounds for the second term that are tighter than the usual, $(R_j^{max} - R_j^{min}) T$.

Consequently, the total error in predicting $j$'s behavior is bounded if lower-level models are solved exactly. Otherwise, as we show in Section 6.2, the error is large only for very large $\epsilon$. This is because many problems admit large BE regions for the models thereby not overconstraining $\epsilon$, and the prediction continues to remain exact. However, we noticed that these regions do reduce in size as the horizon increases. In summary, although the error due to associating different models whose beliefs are $\epsilon$-close is bounded, we are unable to usefully bound the overall error in prediction due to approximate solutions of lower-level models.

## Appendix B. Problem Domains

We provide detailed descriptions of all the problem domains utilized in our evaluations, including their I-DID models, below.

### B.1 Multiagent Tiger Problem

As we mentioned previously, our multiagent tiger problem is a non-cooperative generalization of the well-known single agent tiger problem (Kaelbling et al., 1998) to the multiagent setting. It differs from other multiagent versions of the same problem (Nair et al., 2003) by assuming that the agents hear creaks as well as the growls and the reward function does not promote cooperation. Creaks are indicative of which door was opened by the other agent(s). While we described the problem in Section 3, we quantify the different uncertainties here. We assume that the accuracy of creaks is 90%, while the accuracy of growls is 85% as in the single agent problem. The tiger location is chosen randomly in the next time step if any of the agents opened any doors in the current step.

Fig. 8 shows an I-DID unrolled over two time-slices for the multiagent tiger problem. We give the CPTs for the different nodes below:

| $\langle a_i^t, a_j^t \rangle$ | $\textbf{TigerLocation}^t$ | **TL** | **TR** |
|---|---|---|---|
| $\langle OL, * \rangle$ | * | 0.5 | 0.5 |
| $\langle OR, * \rangle$ | * | 0.5 | 0.5 |
| $\langle *, OL \rangle$ | * | 0.5 | 0.5 |
| $\langle *, OR \rangle$ | * | 0.5 | 0.5 |
| $\langle L, L \rangle$ | $TL$ | 1.0 | 0 |
| $\langle L, L \rangle$ | $TR$ | 0 | 1.0 |

Table 2: CPT of the chance node $TigerLocation^{t+1}$ in the I-DID of Fig. 8.

We assign the marginal distribution over the tiger's location from agent $i$'s initial belief to the chance node, $TigerLocation^t$. The CPT of $TigerLocation^{t+1}$ in the next time step conditioned on $TigerLocation^t$, $A_i^t$, and $A_j^t$ is the transition function, shown in Table 2. The CPT of the observation node, $Growl\&Creak^{t+1}$, is shown in Table 3. CPTs of the observation nodes in level 0 DIDs are identical to the observation function in the single agent tiger problem.

| $\langle a_i^t, a_j^t \rangle$ | $\textbf{TgrLoc}^{t+1}$ | $\langle \textbf{GL, CL} \rangle$ | $\langle \textbf{GL, CR} \rangle$ | $\langle \textbf{GL, S} \rangle$ | $\langle \textbf{GR, CL} \rangle$ | $\langle \textbf{GR, CR} \rangle$ | $\langle \textbf{GR, S} \rangle$ |
|---|---|---|---|---|---|---|---|
| $\langle L, L \rangle$ | $TL$ | 0.85*0.05 | 0.85*0.05 | 0.85*0.9 | 0.15*0.05 | 0.15*0.05 | 0.15*0.9 |
| $\langle L, L \rangle$ | $TR$ | 0.15*0.05 | 0.15*0.05 | 0.15*0.9 | 0.85*0.05 | 0.85*0.05 | 0.85*0.9 |
| $\langle L, OL \rangle$ | $TL$ | 0.85*0.9 | 0.85*0.05 | 0.85*0.05 | 0.15*0.9 | 0.15*0.05 | 0.15*0.05 |
| $\langle L, OL \rangle$ | $TR$ | 0.15*0.9 | 0.15*0.05 | 0.15*0.05 | 0.85*0.9 | 0.85*0.05 | 0.85*0.05 |
| $\langle L, OR \rangle$ | $TL$ | 0.85*0.05 | 0.85*0.9 | 0.85*0.05 | 0.15*0.05 | 0.15*0.9 | 0.15*0.05 |
| $\langle L, OR \rangle$ | $TR$ | 0.15*0.05 | 0.15*0.9 | 0.15*0.05 | 0.85*0.05 | 0.85*0.9 | 0.85*0.05 |
| $\langle OL, * \rangle$ | * | 1/6 | 1/6 | 1/6 | 1/6 | 1/6 | 1/6 |
| $\langle OR, * \rangle$ | * | 1/6 | 1/6 | 1/6 | 1/6 | 1/6 | 1/6 |

Table 3: CPT of the chance node, $Growl\&Creak^{t+1}$, in agent $i$'s I-DID.

Decision nodes, $A_i^t$ and $A_i^{t+1}$, contain possible actions of agent $i$ such as *L*, *OL*, and *OR*. Model node, $M_{j,l-1}^t$, contains the different models of agent $j$ which are DIDs if the I-DID is at level 0, otherwise they are I-DIDs themselves. The distribution over the associated $Mod[M_j^t]$ node (see Fig. 9) is the conditional distribution over $j$'s models given physical state from agent $i$'s initial belief. The CPT of the chance node, $Mod[M_j^{t+1}]$, in the model node, $M_{j,l-1}^t$, reflects which prior model, action and observation of $j$ results in a model contained in the model node.

Finally, the utility node, $R_i$, in the I-DID relies on both agents' actions, $A_i^t$ and $A_j^t$, and the physical states, $TigerLocation^t$. The utility table is shown in Table 4. These payoffs are analogous to the single agent version, which assigns a reward of 10 if the correct door is opened, a penalty of 100 if the opened door is the one behind which is a tiger, and a penalty of 1 for listening. A result of this assumption is that the other agent's actions do not impact the original agent's payoffs directly, but rather indirectly by resulting in states that matter to the original agent. The utility tables for level 0 models are exactly identical to the reward function in the single agent tiger problem.

| $\langle a_i, a_j \rangle$ | **TL** | **TR** |
|---|---|---|
| $\langle OR, OR \rangle$ | 10 | -100 |
| $\langle OL, OL \rangle$ | -100 | 10 |
| $\langle OR, OL \rangle$ | 10 | -100 |
| $\langle OL, OR \rangle$ | -100 | 10 |
| $\langle L, L \rangle$ | -1 | -1 |
| $\langle L, OR \rangle$ | -1 | -1 |
| $\langle OR, L \rangle$ | 10 | -100 |
| $\langle L, OL \rangle$ | -1 | -1 |
| $\langle OL, L \rangle$ | -100 | 10 |

Table 4: Utility table for node, $R_i$, in the I-DID. Utility table in the I-DID for agent $j$ is the same with column label, $\langle a_i, a_j \rangle$, swapped.

## B.2 Multiagent Machine Maintenance Problem

We extend the traditional single agent based machine maintenance (MM) problem (Smallwood & Sondik, 1973) to a two-agent cooperative version. Smallwood and Sondik (1973) described an MM problem involving a machine containing two internal components. Either one or both components of the machine may fail spontaneously after each production cycle (0-fail: no component fails; 1-fail: 1 component fails; 2-fail: both components fail). If an internal component has failed, then there is some chance that when operating upon the product, it will cause the product to be defective. An agent may choose to manufacture the product (M) without examining it, examine the product (E), inspect the machine (I), or repair it (R) before the next production cycle. On an examination of the product, the subject may find it to be defective. Of course, if more components have failed, then the probability that the product is defective is greater.

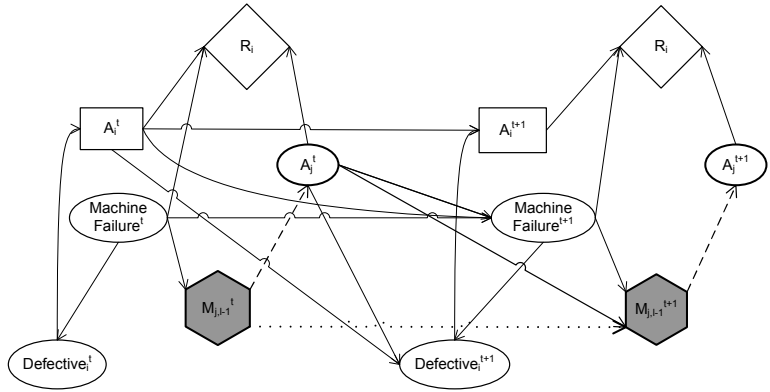

Figure 28: Level $l$ I-DID of agent $i$ for the multiagent MM problem.

A level $l$ I-DID for the multiagent MM problem is shown in Fig. 28. We consider $M$ models of agent $j$ at the lower level which differ in the probability that $j$ assigns to the chance node *Machine Failure*. Agent $i$'s initial belief over the physical state and $j$'s models provides the marginal distribution over $MachineFailure^t$. In the I-DID, the chance node, $MachineFailure^{t+1}$, has incident arcs from the nodes $MachineFailure^t$, $A_i^t$, and $A_j^t$. In Table 5, we show the CPT of the chance node.

| $\langle a_i^t, a_j^t \rangle$ | **Machine Failure**$^{t+1}$ | **0-fail** | **1-fail** | **2-fail** |
|---|---|---|---|---|
| $\langle$M/E,M/E$\rangle$ | 0-fail | 0.81 | 0.18 | 0.01 |
| $\langle$M/E,M/E$\rangle$ | 1-fail | 0.0 | 0.9 | 0.1 |
| $\langle$M/E,M/E$\rangle$ | 2-fail | 0.0 | 0.0 | 1.0 |
| $\langle$M,I/R$\rangle$ | 0-fail | 1.0 | 0.0 | 0.0 |
| $\langle$M,I/R$\rangle$ | 1-fail | 0.95 | 0.05 | 0.0 |
| $\langle$M,I/R$\rangle$ | 2-fail | 0.95 | 0.0 | 0.05 |
| $\langle$E,I/R$\rangle$ | 0-fail | 1.0 | 0.0 | 0.0 |
| $\langle$E,I/R$\rangle$ | 1-fail | 0.95 | 0.05 | 0.0 |
| $\langle$E,I/R$\rangle$ | 2-fail | 0.95 | 0.0 | 0.05 |
| $\langle$I/R,*$\rangle$ | 0-fail | 1.0 | 0.0 | 0.0 |
| $\langle$I/R,*$\rangle$ | 1-fail | 0.95 | 0.05 | 0.0 |
| $\langle$I/R,*$\rangle$ | 2-fail | 0.95 | 0.0 | 0.05 |

Table 5: CPT of the chance node, $MachineFailure^{t+1}$, in the level $l$ I-DID of agent $i$. At level 0 the CPT is analogous to the one in the original MM problem.

With the observation chance node, $Defective_i^{t+1}$, we associate the CPT shown in Table 6. Note that arcs from $MachineFailure^{t+1}$ and the nodes, $A_i^t$ and $A_j^t$, in the previous time step are incident to this node. The observation nodes in the level 0 DIDs have CPTs that are identical to the observation function in the original MM problem.

| $\langle a_i^t, a_j^t \rangle$ | **Machine Failure**$^{t+1}$ | **not-defective** | **defective** |
|---|---|---|---|
| $\langle$M,M/E$\rangle$ | * | 0.5 | 0.5 |
| $\langle$M,I/R$\rangle$ | * | 0.95 | 0.05 |
| $\langle$E,M/E$\rangle$ | 0-fail | 0.75 | 0.25 |
| $\langle$E,M/E$\rangle$ | 1-fail | 0.5 | 0.5 |
| $\langle$E,M/E$\rangle$ | 2-fail | 0.25 | 0.75 |
| $\langle$E,I/R$\rangle$ | * | 0.95 | 0.05 |
| $\langle$I/R,*$\rangle$ | * | 0.95 | 0.05 |

Table 6: CPT of the observation node, $Defective_i^{t+1}$. Corresponding CPT in agent $j$'s $l-1$ I-DID is identical but with $\langle a_i^t, a_j^t \rangle$ swapped.

The decision node, $A_i$, has one information arc from the observation node $Defective_i^t$ indicating that $i$ knows the examination results before making the choice. The utility node $R_i$ is associated with the utility table in Table 7. The utility table for a level 0 agent is identical to the one in the original MM problem.

The CPT of the chance node, $Mod[M_j^{t+1}]$, in the model node, $M_{j,l-1}^{t+1}$, reflects which prior model, action and observation of $j$ results in a model contained in the model node, analogously to the tiger problem.

| $\langle a_i^t, a_j^t \rangle$ | 0-fail | 1-fail | 2-fail |
|---|---|---|---|
| $\langle$M,M$\rangle$ | 1.805 | 0.95 | 0.5 |
| $\langle$M,E$\rangle$ | 1.555 | 0.7 | 0.25 |
| $\langle$M,I$\rangle$ | 0.4025 | -1.025 | -2.25 |
| $\langle$M,R$\rangle$ | -1.0975 | -1.525 | -1.75 |
| $\langle$E,M$\rangle$ | 1.5555 | 0.7 | 0.25 |
| $\langle$E,E$\rangle$ | 1.305 | 0.45 | 0.0 |
| $\langle$E,I$\rangle$ | 0.1525 | -1.275 | -2.5 |
| $\langle$E,R$\rangle$ | -1.3475 | -1.775 | -2.0 |
| $\langle$I,M$\rangle$ | 0.4025 | -1.025 | -2.25 |
| $\langle$I,E$\rangle$ | 0.1525 | -1.275 | -2.5 |
| $\langle$I,I$\rangle$ | -1.0 | -3.00 | -5.00 |
| $\langle$I,R$\rangle$ | -2.5 | -3.5 | -4.5 |
| $\langle$R,M$\rangle$ | -1.0975 | -1.525 | -1.75 |
| $\langle$R,E$\rangle$ | -1.3475 | -1.775 | -2.0 |
| $\langle$R,I$\rangle$ | -2.5 | -3.5 | -4.5 |
| $\langle$R,R$\rangle$ | -4 | -4 | -4 |

Table 7: Utility table for agent $i$. Agent $j$'s utility table in its $l-1$ I-DID is identical but with column label, $\langle a_i^t, a_j^t \rangle$, swapped.

### B.3 UAV Reconnaissance Problem

We show a level $l$ I-DID for the multiagent UAV problem in Fig 29. Models of the fugitive (agent $j$) at the lower level differ in the probability that the fugitive assigns to its position in the grid. The UAV's (agent $i$) initial beliefs are probability distributions assigned to the relative position of the fugitive decomposed into the chance nodes, $FugRelPosX^t$ and $FugRelPosY^t$, which represent the relative location of the fugitive along the row and column, respectively. Its CPTs assume that each action (except *listen*) moves the UAV in the intended direction with a probability of 0.67, while the remaining probability is equally divided among the other neighboring positions. Action *listen* keeps the UAV in the same position.

The observation node, $SenFug$, represents the UAV's sensing of the relative position of the fugitive in the grid. Its CPT assumes that the UAV has good sensing capability (likelihood of 0.8 for the correct relative location of fugitive) if the action is *listen*, otherwise the UAV receives random observations during other actions.

The decision node, $A_i$, contains five actions of the UAV, which includes moving in the four cardinal directions and listening. The edge incident into the node indicates that the UAV ascertains the observation on the relative position of the fugitive before it takes an action.

The utility node, $R_i$, is the reward assigned to the UAV for its actions given the fugitive's relative position and its actions. The UAV gets rewarded 50 if it captures the fugitive; otherwise, it costs -5 for performing any other action.

Because the actual CPT tables are very large, we do not show them here. All problem domain files are available upon request.

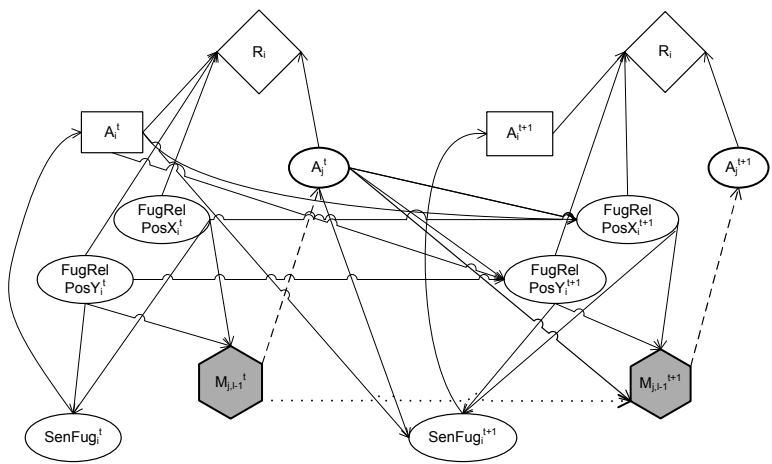

Figure 29: Level $l$ I-DID of agent $i$ for our UAV reconnaissance problem.